\documentclass[sigconf]{acmart}

\AtBeginDocument{%
  }

\setcopyright{acmlicensed}
\setcopyright{none}
\copyrightyear{2025}
\acmYear{2025}
\acmDOI{XXXXXXX.XXXXXXX}

\renewcommand\footnotetextcopyrightpermission[1]{}

\acmConference[Conference acronym 'XX]{Make sure to enter the correct
 conference title from your rights confirmation emai}{June 03--05,
 2018}{Woodstock, NY}
\acmBooktitle{Woodstock '18: ACM Symposium on Neural Gaze Detection,
 June 03--05, 2018, Woodstock, NY}
\acmISBN{978-1-4503-XXXX-X/18/06}

\usepackage{amssymb}
\usepackage{amsmath}
\usepackage{colortbl}
\usepackage{xcolor}
\usepackage{algorithm2e}
\usepackage{multirow}
\usepackage{subfigure}
\usepackage{threeparttable}
\usepackage{tablefootnote}
\usepackage{subcaption} 
\usepackage{tabularx}
\usepackage{longtable}
\usepackage{xltabular}
\usepackage{diagbox}
\usepackage{ltxtable}
\usepackage{graphicx} 
\usepackage{array} 
\DeclareMathOperator*{\Motimes}{\text{\raisebox{0.3ex}{\scalebox{0.7}{$\bigotimes$}}}}

\DeclareMathOperator*{\Moplus}{\text{\raisebox{0.3ex}{\scalebox{0.7}{$\bigoplus$}}}}

\newcolumntype{Y}{>{\raggedright\arraybackslash}X}
\usepackage[most]{tcolorbox}
\usepackage{tcolorbox}
\definecolor{mybackcolor}{RGB}{240, 246, 232}
\definecolor{myframecolor}{RGB}{102,166,30}
\definecolor{mybackcolor1}{RGB}{242,242,242}
\definecolor{myframecolor1}{RGB}{219,219,219}
\usepackage{lipsum} 
\tcbset{
    mytakeaway/.style={
        enhanced,
        breakable,  
        colback=mybackcolor,
        colframe=myframecolor,
        left=2mm,
        right=2mm,
        boxrule=0mm,
        arc=0mm,
        width=\columnwidth,
        before skip=10pt,
        after skip=10pt,
        frame hidden, 
        borderline west={1mm}{0mm}{myframecolor}, 
    }
}

\tcbuselibrary{listings} 

\newtheorem{prompt}{Prompt}
\tcbset{
    prompt/.style={
        enhanced,
        breakable,  
        colback=mybackcolor1,
        colframe=myframecolor1,
        left=2mm,
        right=2mm,
        boxrule=0mm,
        arc=0mm,
        before skip=10pt,
        after skip=10pt,
        frame hidden, 
        borderline west={0mm}{0mm}{myframecolor1}, 
        title=#1, 
        coltitle=black, 
        fonttitle=\bfseries, 
        colbacktitle=myframecolor1, 
        width=\columnwidth, 
        left=0pt, 
        right=0pt, 
    }
}

\usepackage{tabularray}
\UseTblrLibrary{booktabs,siunitx}

\ifodd 1

\else

\fi

\ifodd 1

\else

\fi

\begin{document}


\title{Way to Specialist: Closing Loop Between Specialized LLM and Evolving Domain Knowledge Graph} 


\author{Yutong Zhang}
\orcid{0009-0006-4959-2774}
\affiliation{%
\institution{Shanghai Jiao Tong University}
\department{School of Cyber Science and Engineering}
  \streetaddress{}
  \city{Shanghai}
  \state{}
  \country{China}
  \postcode{200240}
}
\email{zhangyutong.sjtu@sjtu.edu.cn}

\author{Lixing Chen}
\orcid{0000-0002-1805-0183}
\authornotemark[1]
\affiliation{%
  \institution{Shanghai Jiao Tong University}
  \department{Institute of Cyber Science and Engineering}
  \streetaddress{}
  \city{Shanghai}
  \state{}
  \country{China}
  \postcode{200240}
}
\email{lxchen@sjtu.edu.cn}

\author{Shenghong Li}
\orcid{0000-0002-0767-2307}
\affiliation{%
  \institution{Shanghai Jiao Tong University}
  \department{School of Cyber Science and Engineering}
  \streetaddress{}
  \city{Shanghai}
  \state{}
  \country{China}
  \postcode{200240}
}
\email{shli@sjtu.edu.cn}
\authornote{Corresponding Author}

\author{Nan Cao}
\orcid{0000-0003-1316-7515}
\affiliation{%
  \institution{Tongji University}
  \department{College of Design and Innovation}
  \streetaddress{}
  \city{Shanghai}
  \state{}
  \country{China}
  \postcode{200092}
}
\email{nan.cao@tongji.edu.cn}

\author{Yang Shi}
\orcid{0000-0002-1065-4038}
\affiliation{%
  \institution{Tongji University}
  \department{College of Design and Innovation}
  \streetaddress{}
  \city{Shanghai}
  \state{}
  \country{China}
  \postcode{200092}
}
\email{yangshi.idvx@tongji.edu.cn}

\author{Jiaxin Ding}
\orcid{0000-0002-0009-9237}
\affiliation{%
  \institution{Shanghai Jiao Tong University}
  \department{John Hopcroft Center for Computer Science}
  \streetaddress{}
  \city{Shanghai}
  \state{}
  \country{China}
  \postcode{200240}
}
\email{jiaxinding@sjtu.edu.cn}

\author{Zhe Qu}
\orcid{0000-0003-2211-2137}
\affiliation{%
  \institution{Central South University}
  \department{School of Computer Science and Engineering}
  \streetaddress{}
  \city{Changsha}
  \state{Hunan}
  \country{China}
  \postcode{410083}
}
\email{zhe_qu@csu.edu.cn}

\author{Pan Zhou}
\orcid{0000-0002-8629-4622}
\affiliation{%
  \institution{Huazhong University of Science and Technology}
  \department{School of Cyber Science and Engineering}
  \streetaddress{}
  \city{Wuhan}
  \state{Hubei}
  \country{China}
  \postcode{430070}
}
\email{panzhou@hust.edu.cn}

\author{Yang Bai}
\orcid{0000-0002-1037-3973}
\affiliation{%
  \institution{Shanghai Jiao Tong University}
  \department{Department of Automation}
  \streetaddress{}
  \city{Shanghai}
  \state{}
  \country{China}
  \postcode{200240}
}
\email{ybai@sjtu.edu.cn}

\begin{abstract}
Large language models (LLMs) have demonstrated exceptional performance across a wide variety of domains. Nonetheless, generalist LLMs continue to fall short in reasoning tasks necessitating specialized knowledge. Prior investigations into specialized LLMs focused on domain-specific training, which entails substantial efforts in domain data acquisition and model parameter fine-tuning. To address these challenges, this paper proposes the Way-to-Specialist (WTS) framework, which synergizes retrieval-augmented generation with knowledge graphs (KGs) to enhance the specialized capability of LLMs in the absence of specialized training. In distinction to existing paradigms that merely utilize external knowledge from general KGs or static domain KGs to prompt LLM for enhanced domain-specific reasoning, WTS proposes an innovative "LLM$\circlearrowright$KG" paradigm, which achieves bidirectional enhancement between specialized LLM and domain knowledge graph (DKG). The proposed paradigm encompasses two closely coupled components: the \emph{DKG-Augmented LLM} and the \emph{LLM-Assisted DKG Evolution}. The former retrieves question-relevant domain knowledge from DKG and uses it to prompt LLM to enhance the reasoning capability for domain-specific tasks; the latter leverages LLM to generate new domain knowledge from processed tasks and use it to evolve DKG. WTS closes the loop between \emph{DKG-Augmented LLM} and \emph{LLM-Assisted DKG Evolution}, enabling continuous improvement in the domain specialization as it progressively answers and learns from domain-specific questions. We validate the performance of WTS on 6 datasets spanning 5 domains. The experimental results show that WTS surpasses the previous SOTA in 4 specialized domains and achieves a maximum performance improvement of 11.3\%. 
\end{abstract}



\maketitle

\section{Introduction}

Large language models (LLMs), e.g., GPT-4 \cite{achiam2023gpt}, Gemini \cite{team2023gemini}, and Llama \cite{touvron2023llama} have demonstrated their exceptional performance across a wide range of general domains \cite{ouyang2022training, achiam2023gpt, thoppilan2022lamda, NEURIPS2020_1457c0d6, chowdhery2023palm, touvron2023llama}. Yet, there is a prevalent recognition that LLMs often exhibit limitations when confronted with reasoning tasks that necessitate specialized knowledge \cite{peng2023check}. Most explorations of LLMs to date for specific domains, e.g., medical \cite{yang2024fine,liu2023tailoring} and law \cite{sun2024lawluo} fields, have leveraged domain-specialized training or parameter fine-tuning techniques in pursuit of performance enhancement. These processes necessitate the acquisition of high-quality instruction data and sophisticated design of training pipelines, both of which entail substantial efforts \cite{cao2023instruction}.

In light of these challenges, we focus on steering foundation models via Retrieval-Augmented Generation (RAG) \cite{shuster2021retrieval,mallen2022not} to excel in specialty areas. RAG falls under the category of prompt engineering technique \cite{wei2022chain, NEURIPS2020_1457c0d6, yao2024tree}. It uses a \emph{question-retrieve-generate} framework --- Given a domain-specific question, external knowledge relevant to the question is retrieved from a knowledge source, which is then incorporated in input prompts to instruct LLM for answer generation. RAG works without modifications to model parameters, thereby offering cost-effective and rapid implementations. However, empirical evidence indicates that these methods constantly provide disturbing information, causing the knowledge noise issue \cite{liu2020k} to RAG-based LLMs. Therefore, state-of-the-art (SOTA) RAG approaches \cite{baek2023knowledge, sun2023think, wen2023mindmap} are shifting towards the use of knowledge graphs (KG) as knowledge sources (termed as KG-augmented LLM), which offers a structured and explicit representation of knowledge and facilitates the retrieval of high-quality knowledge relevant to questions.

\begin{figure*}[htpb]
\includegraphics[width=\linewidth]{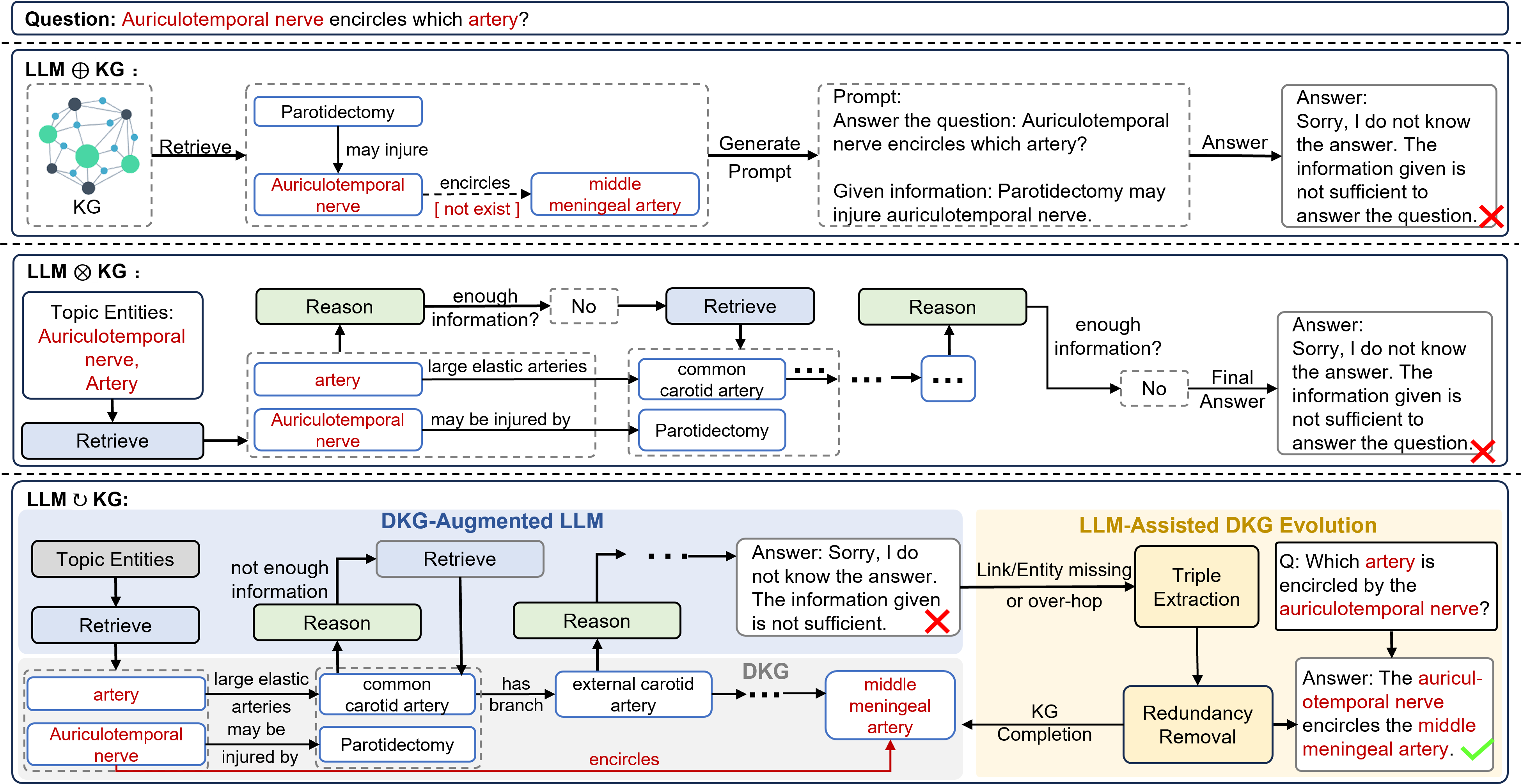}
\vspace{-0.2 in}
  \caption{Comparison of ``LLM$\circlearrowright$KG'' against SOTA paradigms in KG-augmented LLM. }
  \vspace{-0.15 in}
  \label{fig:intro}
\end{figure*}

Current KG-augmented LLMs methods mainly utilize paradigms of ``LLM$\Moplus$KG'' \cite{baek2023knowledge,xie2022unifiedskg} and ``LLM$\Motimes$KG'' \cite{sun2023think,wen2023mindmap}. As illustrated in Figure \ref{fig:intro}, the ``LLM$\Moplus$KG'' paradigm follows a systematic procedure: it retrieves information from KGs to augment the prompt and then inputs the enhanced prompt into LLMs. 
The "LLM$\Motimes$KG" paradigm improves "LLM$\Moplus$KG" framework by incorporating sophisticated interactions between KGs and large LLMs, which partially addresses the challenges associated with multi-hop reasoning. These works \cite{wu2023retrieve, sun2023think, luo2023reasoning} utilize general KGs, e.g., Wikidata \cite{vrandevcic2014wikidata}, Freebase \cite{bollacker2008freebase}, to enhance their performance for common-sense questions. Nevertheless, the knowledge encapsulated in general KGs is coarse and often lacks specialized information, potentially leading to knowledge mismatches when addressing domain-specific questions. Due to this limitation, recent advancements \cite{wen2023mindmap,soman2023biomedical,matsumoto2024kragen,taffa2023leveraging} have been inspired to incorporate Domain Knowledge Graphs (DKGs) to KG-augmented LLMs to enhance domain-specific reasoning, particularly in the medical field \cite{wen2023mindmap,soman2023biomedical,matsumoto2024kragen}. However, these studies rest on the precondition that a comprehensive DKG has been established already, which is currently true only for a few specialized domains, e.g., medical \cite{wen2023mindmap, soman2023biomedical, matsumoto2024kragen} and food knowledge \cite{haussmann2019foodkg}. Most domains continue to face challenges with the absence and incompleteness of DKGs. 
The utilization of low-quality DKGs can adversely affect the performance of domain-specific reasoning. Furthermore, these SOTA solutions utilize static DKGs for knowledge retrieval and prompt generation, which may run the risk of being outdated and misaligned with the evolving knowledge demand during implementation \cite{xu2024generate}. Additionally, DKGs may also require personalization or customization to address unique needs, which is also overlooked in previous works. These challenges underscore the necessity of employing evolving DKGs that continuously update and integrate appropriate domain knowledge.

To fill in the above deficiencies, this work introduces the Way-to-Specialist (WTS) framework, which designs an innovative strategy for developing specialized LLMs through the mutual enhancement between LLM and DKG. In contrast to the paradigms ``LLM$\Moplus$KG'' and ``LLM$\Motimes$KG'' that operate in unidirectional enhancement processes using KG to augment input prompts of LLMs, WTS designs a novel ``LLM$\circlearrowright$KG'' paradigm which not only leverages DKG to prompt LLM for the enhancement of domain-specific reasoning capability but also utilizes LLM to evolve DKG for the enhancement of domain knowledge comprehensiveness (illustrated in Figure \ref{fig:intro}). Such bidirectional enhancement between LLM and DKG forms interactive loops as WTS progressively addresses and learns from domain-specific questions. The novelties and contributions of our work are summarized as follows:

1) WTS introduces an innovative LLM$\circlearrowright$KG paradigm which enables the bidirectional enhancement of LLM and DKG for constructing specialized LLMs. This paradigm comprises two integrated components: the \emph{DKG-Augmented LLM} and the \emph{LLM-Assisted DKG Evolution}. The \emph{DKG-Augmented LLM} retrieves question-relevant domain knowledge from the DKG and uses it to prompt LLM for answer generation. Subsequently, the \emph{LLM-Assisted DKG Evolution} leverages LLM to generate new knowledge from the processed question and use it to evolve the DKG. 
In particular, WTS can initiate with an empty or incomplete DKG and progressively complete it to enhance the reasoning capability for domain-specific tasks.

2) WTS designs an interactive knowledge retrieval mechanism for \emph{DKG-Augmented LLM}. WTS maintains DKGs in the form of vector databases. In each retrieval iteration, WTS assesses the similarity of questions and knowledge in DKG by measuring the distance between their feature embeddings generated by the language embedding model of the vector database. Further, WTS utilizes LLM to evaluate the semantic relevance of knowledge triples to the question and prune less relevant ones. The retrieved knowledge is used to prompt the question reasoning, and the LLM generates the answer along with a recommendation to determine whether the retrieval proceeds to the next iteration to further enrich the external knowledge.  

3) WTS proposes an LLM-based knowledge generation scheme and a redundancy-aware DKG update mechanism for \emph{LLM-Assisted DKG Evolution}. LLM-based knowledge generation leverages LLM to efficiently generate knowledge triples from the unstructured information in questions, answers, and knowledge retrieved by \emph{DKG-Augmented LLM}. 
The redundancy-aware DKG update mechanism evaluates the redundancy between newly generated knowledge and existing knowledge in DKG, then excludes redundant knowledge during evolution to guarantee the knowledge efficiency of DKG, which facilitates the knowledge retrieval in \emph{DKG-Augmented LLM}.

4) We conduct experiments of WTS on 6 datasets spanning 5 domains, including 4 specialized domains (medical, natural science, social science, and linguistics) and 1 general domain. The results show that WTS achieves overall the best performance in all 4 specialized domains, 
achieving a maximum performance improvement of 11.3\% compared to the current SOTA.

\section{Related Works}

\indent \textbf{Large Language Models \& Knowledge Graphs:}
The integration of KG and LLM can be categorized into two branches: KG for LLM and LLM for KG. The works in the branch of \underline{\textbf{KG for LLM}} mainly use Retrieval-Augmented Generation (RAG) \cite{lewis2020retrieval} to incorporate non-parametric information \cite{wang2019improving, saxena2020improving, sun2020colake, liu2020k, zhang2022drlk} from KGs to pre-trained LLM for addressing knowledge-intensive text generation task. 
KAPING \cite{baek2023knowledge} pioneered KG-augmented LLM by retrieving relevant knowledge triples from KGs and prepending them to the input question as a prompt. KAPING only utilizes a single-layer KG, making it challenging to address questions that require multi-hop knowledge analysis on KGs. To address this issue, the works \cite{wu2023retrieve, sun2023think, luo2023reasoning, agarwal2023bring} conduct multi-hop knowledge retrieval over KGs to improve the reasoning capacity of LLMs. KG-augmented LLMs also have spurred research into specialized LLMs in medical field \cite{wen2023mindmap, jiang2023think,soman2023biomedical,matsumoto2024kragen}, politics \cite{mou2024unifying}, scholar \cite{taffa2023leveraging} and law \cite{cui2023chatlaw}. These works utilize existing RAG frameworks, e.g. mind map \cite{wen2023mindmap}, reason chain \cite{jiang2023think}, and aligned embeddings \cite{mou2024unifying}, to generate input prompts. The works in the branch of \underline{\textbf{LLM for KG}} leverages LLMs to support knowledge engineering tasks \cite{meyer2023llm}, including KG construction \cite{carta2023iterative, bi2024codekgc} and KG completion \cite{sadeghian2019drum,yao2019kg, wang2022simkgc}.
KG construction involves collecting and integrating information from various sources \cite{pan2023large}. Recent works have adopted both schema-based methods \cite{carta2023iterative, hu2023llm, mihindukulasooriya2023text2kgbench} and schema-free methods \cite{bi2024codekgc} with LLM to construct KGs based on structured texts. KG completion involves inferring missing knowledge triples. These works are typically categorized into rule-based methods \cite{yang2017differentiable, sadeghian2019drum}, embedding-based methods \cite{bordes2013translating, trouillon2016complex}, and text-based methods \cite{yao2019kg, wang2022simkgc, wang2021structure}. 
LLM is mostly implemented with text-based methods due to its exceptional capability in semantic understanding. For example, KICGPT \cite{wei2024kicgpt} and KERMIT \cite{li2023kermit} utilized LLMs as entity-sorting assistants and data-enhancement tools and validated the advantage of using LLMs in KG completion. However, these methods were limited to completing the KG based on existing entities and could not introduce new entities and triples into the KG.

\textbf{Specialized LLM:} Despite the remarkable performance of LLMs, they still struggle with problems that require specialized domain expertise \cite{mialon2023augmented}. Recent efforts have sought to address this issue through fine-tuning, which involves partial modification of model parameters, or through RAG-based prompting, which incorporates specialized domain knowledge, e.g., in the medical \cite{xia2022medconqa, wen2023mindmap, jiang2023think, soman2023biomedical, matsumoto2024kragen} and finance \cite{baldazzi2023fine} field.
The works \cite{thirunavukarasu2023large, jiang2024efficient, liu2024moe, yang2024fine, baldazzi2023fine} have investigated fine-tuning techniques to LLMs and achieved considerable performance improvements for domain-specific questions. However, fine-tuning techniques require substantial computational resources and time and may lead to catastrophic forgetting \cite{lin2023speciality} where the LLM loses its general knowledge while adapting to the specialized domain. Compared to fine-tuning techniques, RAG dynamically integrates up-to-date external domain knowledge into LLM reasoning, providing enhanced efficiency and simplified implementation. These works, e.g., MVPKG in \cite{mou2024unifying} and CMCKG \cite{wen2023mindmap}, all utilize manually constructed KGs. However, most specialized domains still lack available KGs, which hinders the implementation of KG-augmented LLM. Furthermore, KGs may become outdated, potentially diminishing the effectiveness of RAG methods.

\section{Methodology of Way-to-Specialist}
\subsection{Preliminaries}
As WTS rests on a synergistic integration of domain knowledge graph (DKG) and retrieval-augmented LLM, it is essential to provide a brief introduction to these two techniques prior to delineating the design of WTS.      

\textbf{Domain Knowledge Graph}: A DKG is characterized as a collection of knowledge triples $\mathcal{G} = \{(e_\texttt{s}, r, e_\texttt{o}) | e_\texttt{s} \in \mathcal{E}, r \in \mathcal{R}, e_\texttt{o} \in \mathcal{E}\}$, where $(e_\texttt{s}, r, e_\texttt{o})$ is a knowledge triple, $\mathcal{E}$ is the entity set, $\mathcal{R}$ is the relation set, $e_\texttt{s}$, $r$, and $e_\texttt{o}$ correspond to the subject entity, relation and object entity, respectively. WTS maintains DKG in the form of vector databases\cite{schlegel2019dbee}, where entities and relationships within the DKG are represented as high-dimensional feature embedding. This approach enables efficient storage, retrieval, and analysis of the DKG through vector-based operations, drastically improving the efficiency of DKG-based prompting (will be demonstrated in subsequent sections).

\begin{figure*}[htpb]
\includegraphics[width=\textwidth]{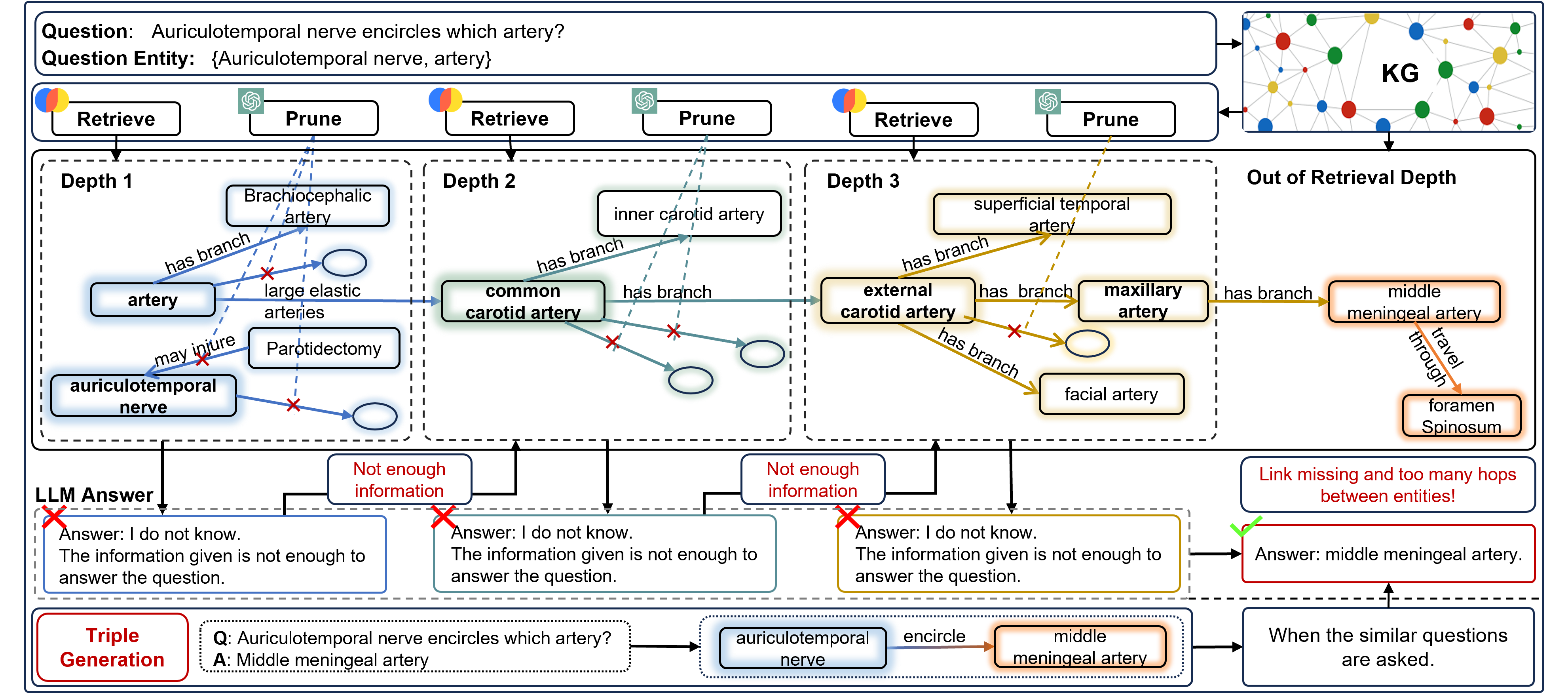}
\vspace{-0.2 in}
  \caption{Overview of WTS. The upper part is \emph{DKG-Augmented LLM} which extracts entities from the received question and performs iterative Retrieval-Prune-Reason processes. 
  The bottom part is \emph{LLM-Assisted DKG Evolution} which generates new knowledge triples from the processed question to evolve the DKG.}
\vspace{-0.15 in}
\label{fig:example}
\end{figure*}

\textbf{Retrieval-augmented LLM}: Retrieval-augmented LLM \cite{baek2023knowledge} involves integrating an external knowledge base with LLM to provide relevant information during the generation process. The key idea is to retrieve pertinent documents or data from an external source and use this information to augment the LLM's predictions. A retrieval-augmented LLM typically comprises two main components: a retriever $\texttt{Retriever}(q) \to \mathcal{I}_q$ to obtain related information about question $q$ and a generator (i.e., LLM) $\texttt{Generator}(q,\mathcal{I}_q) \to \alpha_q$ to produces the answer $\alpha_q$ to question $q$. Retrieval-augmented LLM offers enhanced contextual understanding and mitigates the hallucination problem commonly associated with LLM.

\subsection{Architecture of WTS}
As illustrated in Figure \ref{fig:framework}, WTS encompasses two closely interconnected components: \emph{DKG-Augmented LLM} and \emph{LLM-Assisted DKG Evolution}. \emph{DKG-Augmented LLM} prompts LLM with domain knowledge in DKG to improve the reasoning capability of LLM for domain-specific questions. \emph{LLM-Assisted DKG Evolution} employs LLM to generate knowledge triples from questions and answers, and use them to evolve DKG. 

\subsubsection{\textbf{DKG-Augmented LLM}}
The \emph{DKG-Augmented LLM} consists of four modules: entity extraction module, retrieval module, pruning module, and reasoning module.

\textbf{Entity Extraction Module:} Upon receiving a question $q$, WTS utilizes the entity extraction module to identify the topic entities of the question. While traditional entity extraction methods, e.g., Hidden Markov Models (HMMs) \cite{zhou2002named,morwal2012named} and Conditional Random Fields (CRFs) \cite{patil2020named}, exist for implementation, they often struggle with handling domain terms. 
Inspired by the superior performance of LLMs in entity recognition \cite{wang2023gpt, sancheti2024llm, ashok2023promptner}, we utilize LLM for entity extraction. Specifically, the entity extraction module prompts LLM (the input prompt is given in Appendix \ref{Prompts-WTS}, Prompts \ref{WTS_Question_Entity_Extraction}) to extract $N$ primary entities, denoted by $\mathcal{E}_q =\{e_{q,1}, e_{q,2},...,e_{q,N}\}$, from question $q$. This process can be mathematically written as $\mathcal{E}_q = \texttt{LLM}_\texttt{Ent}(q)$, where $\texttt{LLM}_\texttt{Ent}$ is the prompted LLM. The extracted entities will be fed to the subsequent retrieval module. Note that the number of extracted entities is restricted to $N$ to mitigate the retrieval overhead for non-informative entities. 

\textbf{Retrieval Module:} Given extracted entities, the retrieval module retrieves within the DKG to obtain knowledge triples related to the extracted entities in $\mathcal{E}_q$. For ease of description, we for now consider that the retrieval module operates with a non-empty DKG. Subsequent sections will provide a detailed exposition of the construction and evolution of DKG. The retrieval process executes in an iterative manner, with the retrieval depth progressively increasing over successive iterations. 

We delineate 1st-depth retrieval to facilitate a comprehensive understanding of the retrieval process. The 1st-depth retrieval acquires knowledge triples associated with entities in $\mathcal{E}_q$. It is important to note that DKG can be of considerable scale with dense relations between entities. This complexity presents stressing challenges to retrieval efficiency. Our retrieval module employs a dual approach, using \emph{Exact Match} for coarse-grained retrieval and \emph{Similarity Retrieval} for fine-grained retrieval, to mitigate the retrieval overhead over large DKGs. We let $\mathcal{E}^{(1)}_q$ ($\mathcal{E}^{(1)}_q \gets \mathcal{E}_q$) be the input to the 1st-depth retrieval. For each entity $e \in \mathcal{E}^{(1)}_q$, the retrieval module first executes \emph{Exact Match} as coarse filtering to include knowledge triples in DKG that have $e$ as their subject entity or object entity. Mathematically, the output of \emph{Exact Match} at 1-st depth can be written as $\mathcal{T}^{(1),\texttt{EM}}_q = \{t=(e_s,r,e_o) \in \mathcal{G}~|~e_s = e~\text{or}~e_o = e, \forall e \in \mathcal{E}^{(1)}_q\}$. Subsequently, \emph{Similarity Retrieval} is implemented to refine knowledge triples in $\mathcal{T}^{(1),\texttt{EM}}_q$. \emph{Similarity Retrieval} uses the pre-trained language model, e.g., sentence transformers\footnote{https://huggingface.co/sentence-transformers} or OpenAI embedding models API\footnote{https://platform.openai.com/docs/guides/embeddings}, associated with the vector database to convert question $q$ and knowledge triple $t \in \mathcal{T}^{(1),\texttt{EM}}_q$ to feature embedding. The similarity of triple $t$ and question $q$ is quantified as the cosine distance between their feature embedding, calculated by $S(q,t)=1-\frac{\mathcal{F}(q)\cdot\mathcal{F}(t)}{\|\mathcal{F}(q)\|^2\|\mathcal{F}(t)\|^2}$, where $\mathcal{F}(\cdot)$ denotes the feature embedding operation. A lower $S$ score indicates a higher degree of similarity. We introduce $L$ as the maximum permissible similarity gap, which functions as a threshold to filter out triples that exhibit low similarity to the question. The triples outputted by \emph{Similarity Retrieval} are collected in $\mathcal{T}^{(1), \texttt{SR}}_q = \{t ~|~ S(q,t) \leq L, t \in \mathcal{T}^{(1),\texttt{EM}}_q \}$, which is also the output of 1st-depth retrieval, i.e., $\mathcal{T}^{(1)}_q$.

Upon the completion of the 1st-depth retrieval, the module advances to the next depth to further expand the domain knowledge related to the question. The objective of the 2nd-depth retrieval is to acquire knowledge triples associated with the entities newly identified during the 1st-depth retrieval. The input entities to the 2nd-depth retrieval is $\mathcal{E}^{(2)}_q = \{ e | e \in \mathcal{T}^{(1)}_q, e \notin \mathcal{E}^{(1)}_q\}$. A similar retrieval process will be executed and gives the result of 2nd-depth retrieval $\mathcal{T}^{(2)}_q$. Following this logic, the retrieval module iteratively performs $D$ iterations of triple retrieval, where the input entity set to $d$-depth retrieval is $\mathcal{E}^{(d)}_q = \{ e | e \in \mathcal{T}^{(d-1)}_q, e \notin \bigcup^{d-1}_{i=1}\mathcal{E}^{(i)}_q\}$. The final output of the retrieval module comprises KG triples retrieved over $D$ interactions, which constitutes a sub-graph of DKG. 

\textbf{Pruning Module:} Given the design of the retrieval module, it is evident that the size of the retrieved DKG subgraph expands exponentially as retrieval goes deep. 
The explosion of retrieved knowledge triples not only incurs substantial retrieval overhead but also leads to unfocused and redundant prompts that potentially diminish the performance of specialized LLM. In observation of this issue, WTS introduces a pruning module to identify the most informative knowledge triples to the question.

The pruning module functions during the retrieval process, selectively excluding retrieved triples $\mathcal{T}^{(d)}_q$ at each retrieval depth $d$. At each depth $d$, the pruning module first prompts LLM (the input prompt is given in Appendix \ref{Prompts-WTS}, Prompt \ref{WTS_Triple_Score_and_Prune}) to evaluate the semantic relevance between question $q$ and each knowledge triple $t$ in $\mathcal{T}^{(d)}_{q}$. This evaluation gives a semantic relevance score $M(q,t)\in [0,1]$ for each question-triple pair, mathematically denoted by $M(q,t) = \texttt{LLM}_{\texttt{Sem}}(q,t),~\forall t \in \mathcal{T}^{(d)}_{q}$, where a higher $M$ score indicates greater semantic relevance. 

The pruning module then outputs $K$ triples with the highest semantic relevance score, denoted by $\tilde{\mathcal{T}}^{(d)}_q = \big\{ \mathcal{T} \subseteq \mathcal{T}^{(d)}_q, | \mathcal{T}| = K \mid M(q,t) \geq M(q,t^\prime), \forall t \in \mathcal{T}, \forall t^\prime \in \mathcal{T}^{(d)}_q \backslash \mathcal{T} \big\}$. By applying the pruning module, the retrieved triples at $d$-depth becomes $\tilde{\mathcal{T}}^{(d)}_q$. Correspondingly, the input entity set
to $d$-depth retrieval should be modified as $\mathcal{E}^{(d+1)}_q = \{ e | e \in \mathcal{\tilde{T}}^{(d)}_q, e \notin \bigcup^{d}_{i=1} \mathcal{E}^{(i)}_q\}$.

The above procedures implement width pruning, which constrains the number of knowledge triples at each retrieval depth level. Actually, WTS also executes depth pruning, which entails early exit of retrieval before reaching the maximum depth $D$. The depth pruning is conducted during the reasoning process. Therefore, it will be detailed in the subsequent reasoning module.

\textbf{Reasoning Module:} The reasoning process performs progressively as the retrieval depth increases. Suppose that we have completed $d$-depth of retrieval and pruning, the retrieved knowledge triples are denoted as $\bar{\mathcal{T}}^{(d)}_q = \bigcup^{d}_{i=1} \tilde{\mathcal{T}}^{(i)}_{q}$. The reasoning module arranges the knowledge triples to form a prompt using the format in Appendix \ref{Prompts-WTS}, Prompt \ref{WTS_Reason_with_Triples}, and then feeds the prompt and question to LLM to generate the answer $\alpha_q^{(d)}$ to question $q$, i.e., $\alpha_q^{(d)} = \texttt{LLM}_\texttt{Rea}(q,\bar{\mathcal{T}}^{(d)}_q)$.

The format of the $\alpha^{(d)}_q$ is stipulated with the system prompt \cite{openai_chat_completions} for ease of performance evaluation. 
If the evaluation\cite{sun2023think} is positive, the retrieval stops, otherwise, the retrieval enters the $(d+1)$-th iteration. This process is repeated until the maximum search depth $D$ is reached.

\begin{tcolorbox}[mytakeaway]
    \textbf{Take-away:} \emph{DKG-Augmented LLM} is not merely an isolated sequence of retrieving information from the DKG and reasoning with the LLM. Rather, it involves tight-coupling interactions between DKG and LLM through the process, e.g., LLM is employed for semantic relevance evaluation in retrieval width pruning and early exit judgment in retrieval depth pruning. This allows LLM to retrieve relevant knowledge more efficiently over DKG, thereby enhancing its reasoning capacity for domain-specific questions.
\end{tcolorbox}

\subsubsection{\textbf{LLM-Assisted DKG Evolution}}
LLM-assisted DKG endeavors to enhance the completeness of DKG by progressively incorporating domain knowledge acquired during question answering. The objective of DKG completion is twofold. Firstly, it aims to incorporate new knowledge (entities and relations) previously absent from both DKG and inherent knowledge of LLM. This is pertinent in scenarios where no triples are retrieved for a question or the evaluation of the answer $\alpha^{(D)}_{q}$ remains negative at the final depth $D$. Secondly, DKG completion seeks to enhance the characterization of knowledge dependencies by activating knowledge connections previously neglected. This is pertinent to scenarios where a positive answer is obtained after multiple iterations of knowledge retrieval, which indicates relevant knowledge in the current DKG is not adequately connected. In this case, additional relations should be incorporated into DKG to ensure closer association between related entities.

\textbf{Domain Knowledge Generation:} The fundamental step of DKG evolution is generating knowledge triples based on questions and answers. However, the extraction of knowledge graphs from unstructured data presents significant challenges, primarily due to the inherent variability and complexity of natural language \cite{regino2022natural}. 
Given the irregular and unstructured nature of question-answer pairs, our work leverages LLMs to design a schema-free solution for KG triple generation. Firstly, WTS prompts LLM (the input prompt is given in Appendix \ref{Prompts-WTS}, Prompt \ref{WTS_Generate_KG_Triple}) to extract domain knowledge from question-answer pairs, using entities in the previously retrieved knowledge triples, i.e., $\bar{\mathcal{T}}^{(D)}_q$, as references. Then LLMs generate knowledge triples based on the extracted domain knowledge.
Given question $q$ and the gold answer $\alpha^*_q$, the module prompts LLM to create KG triples $\mathcal{T}^{+}_q = \texttt{LLM}_{\texttt{Gen}}(q, \alpha^*_q, \bar{\mathcal{T}}^{(D)}_q)$. In the case of Figure \ref{fig:example}, the triple generated from this Q\&A pair is \{ subject: auriculotemporal nerve, relation: encircle, object: middle meningeal artery \}.

\textbf{Redundancy Checking and DKG Evolution:} It should be noted that the knowledge triples generated via the above process might already exist in the DKG. The incorporation of these redundant triples would result in increased complexity of the DKG without contributing any new information. Therefore, we perform redundancy checking before incorporating $\mathcal{T}^+_q$ into the DKG. For a generated knowledge triple 
$t=(e_s,r,e_o) \in \mathcal{T}^{+}_q$, WTS first verifies whether the triple $t$ already exists in the DKG, i.e., if ${t}=(e_s,\hat{r},e_o)\in \mathcal{G}$. If it does, the triple will not be added. To further reduce the complexity of the KG, we compare the similarity of $t$ with the existing triples in $\mathcal{G}$ and establish a maximum similarity threshold $L^\prime$. For $\forall t'= (e_s',r',e_o')\in \mathcal{G}$, 
if $S(t',t) \leq L^\prime$ indicating that $t$ and $t'$ convey almost the same semantic information, the new triple $t$ will not be added in $\mathcal{G}$.

\begin{tcolorbox}[mytakeaway]
    \textbf{Take-away:} \emph{LLM-Assisted DKG Evolution} endows a learn-from-experience capability to WTS without parameter refinement/retraining. It enables the DKG to evolve to a refined form as WTS encounters and answers more questions. The evolution of DKG not only enhances its overall comprehensiveness but also allows DKG to achieve a distinctive knowledge distribution that is tailored to the question source.
\end{tcolorbox}

\subsection{WTS Formation Pipeline}
In previous sections, we have presented \emph{DKG-Augmented LLM} and \emph{LLM-Assisted DKG Evolution} independently. Next, to show how these two components work in tandem to complement each other’s capabilities. 

Let us start from the initialization stage. The initial DKG can be either an empty dataset or an established DKG composed of knowledge triples. This enables WTS to support various implementation scenarios, including those where the domain lacks a pre-existing KG or where the available DKG is incomplete. 

\begin{figure}[!htbp]
\vspace{-0.05 in}
\includegraphics[width=0.9\linewidth]{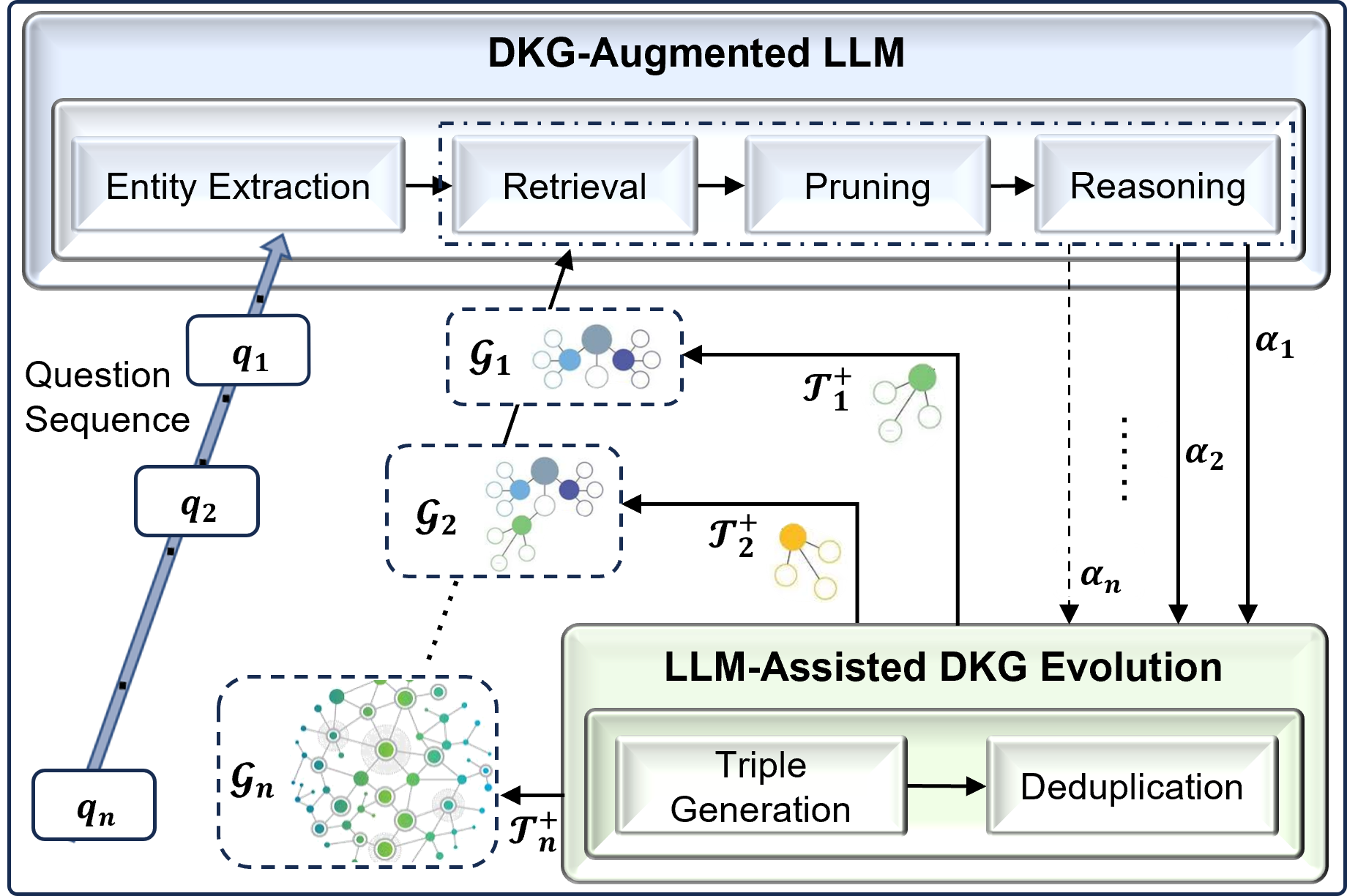}
\vspace{-0.05 in}
  \caption{Illustration of WTS formation pipeline.}
\vspace{-0.2 in}
  \Description{The framework of WTS.}
  \label{fig:framework}
\end{figure}

As shown in Figure \ref{fig:framework}, the formation pipeline of WTS involves cyclical interactions between \emph{DKG-Augmented LLM} and \emph{LLM-Assisted DKG Evolution} throughout the process of question answering. The timeline of WTS formation is discretized by the arrival of questions. The DKG possessed by WTS at the reception of question $q$ is denoted by $\mathcal{G}_{q}$, which is used in DKG-Augmented LLM to prompt the LLM for answer generation. Subsequently, WTS calls \emph{LLM-Assisted DKG Evolution} to generate knowledge triples $\mathcal{T}^{+}_q$ based on question $q$ and gold answer $\alpha^*_q$, and retrieved knowledge triples $\bar{\mathcal{T}}_q^{(D)}$. The generated triples $\mathcal{T}^{+}_q$ are then incorporated to update DKG from $\mathcal{G}_{q}$ to $\mathcal{G}_{q+1}$. The updated DKG $\mathcal{G}_{q+1}$ will be used to answer the subsequent question $q+1$.

From a practical perspective, the formation pipeline of WTS can be split into two phases, Apprenticeship and Mastership, contingent upon the availability of gold answers. During Apprenticeship, WTS functions akin to an apprentice, receiving gold answers from a mentor after addressing questions. This scenario is applicable when comprehensive Q\&A datasets are established in a particular domain. Furthermore, this scenario may also occur when an expert utilizes WTS as an assistant and constantly provides gold answers to questions. The period of Mastership typically follows the Apprenticeship stage, once a DKG has been sufficiently established for prompting domain-specific tasks. During Mastership, WTS operates autonomously without the guidance of gold answers. Instead, WTS relies on the user feedback (e.g., ``Good Response'' and ``Bad Response'' buttons in ChatGPT) to determine whether to extract knowledge triples from generated answers. If positive feedback is received, WTS considers the generated answer as a gold answer to extract $\mathcal{T}^{+}_q$ as aforementioned in \emph{LLM-Assisted DKG Evolution}. Conversely, if negative feedback (or no feedback) is received, WTS relies solely on the question $\mathcal{T}^{+}_q$ for DKG evolution.

\begin{tcolorbox}[mytakeaway]
    \textbf{Take-away:} The formation of WTS is in essence a process of closing the loop between \emph{DKG-Augmented LLM} and \emph{LLM-Assisted DKG Evolution} as WTS progressively answers and learns from domain-specific questions. The connector is an evolving DKG that is utilized by \emph{DKG-Augmented LLM} to prompt the question reasoning and updated by \emph{LLM-Assisted DKG Evolution} based on the retrieved knowledge and answers to enhance domain knowledge comprehensiveness for subsequent questions.
\end{tcolorbox}

\section{Experiments}
\subsection{Experiment Setup}
\subsubsection{Datasets and Baselines}
We evaluate each method on 6 Q\&A datasets, covering 5 domains, including medical  (ChatDoctor5k \cite{li2023chatdoctor}, PubMedQA \cite{jin2019pubmedqa}, MedMCQA \cite{pmlr-v174-pal22a}), natural science, social science, linguistics (SciQ \cite{Welbl2017CrowdsourcingMC},  ScienceQA \cite{saikh2022scienceqa}), and one general domain Q\&A datasets (Simple Questions \cite{bordes2015large}). The MedMCQA dataset includes both single-choice and multiple-choice questions, while in the ScienceQA dataset, SOC, NAT, and LAN refer to the social science, natural science, and linguistics domains, respectively.

For the dataset ChatDoctor5k, we evaluate the text generation task with BERTScore \cite{zhang2019bertscore}, which computes a similarity score of the LLM-generated answer and the real answer. To calculate the BERTScore, we use the bert-base-uncased \cite{devlin2018bert} model as the embedding model. For the other dataset, we use Accuracy as the metric. Detailed information about datasets and metrics is shown in Appendix \ref{Dataset} and \ref{Metrics}.

WTS is compared with 4 SOTA baselines using prompting-based methods, including standard prompting (I/O prompt) \cite{NEURIPS2020_1457c0d6} on ChatGPT-3.5-turbo and GPT-4o, Think-on-Graph (ToG) \cite{sun2023think} with constant Freebase knowledge graph \cite{bollacker2008freebase}, and Chain-of Thought (CoT) \cite{wei2022chain} with 5-shot prompting. 

\subsubsection{Hyperparameters}
The experiments run WTS with two LLMs, ChatGPT-3.5-turbo and GPT-4o, using OpenAI API. The temperature is set to 0.2 and the maximum token length of output is set to 2048. For all datasets, WTS is initialized with an empty vector database as its DKG. With ChromaDB\footnote{https://www.trychroma.com/} serving as the vector database, we employ the pre-trained sentence transformer, all-mpnet-base-v2\footnote{https://huggingface.co/sentence-transformers/all-mpnet-base-v2}, as the embedding model.
The embedding similarity is measured by the cosine distance and the maximum permissible similarity gap is set to $L = 0.55$. To facilitate result evaluation, WTS utilizes prompts to restrict its output format to JSON. This restriction is also applied to GPT-3.5-turbo and GPT-4o baselines for fairness consideration. For the ToG and CoT baselines, minimal instructions are added to their original prompts to ensure consistency in the evaluation process. Further details are provided in Appendix \ref{Evaluation}. 

\subsection{Results and Evaluations}
\subsubsection{Comparison to Baselines}\

\begin{table*}[h]
\vspace{-0.05 in}
\caption{Performance comparison of WTS and baselines on different datasets.}
\vspace{-0.1 in}
\label{tab:main results}
\begin{threeparttable}
\resizebox{\textwidth}{!}{
\begin{tabular}{c|cccccc|cc|c|c|c}
\hline
Domain                         & \multicolumn{6}{c|}{Medical}                                                                                                                             & \multicolumn{2}{c|}{Natural}                                & Social         & Linguistic     & General                   \\ \hline
\multirow{2}{*}{Dataset}       & \multicolumn{3}{c|}{\multirow{2}{*}{ChatDoctor5k}}                    & \multicolumn{1}{c|}{\multirow{2}{*}{PubMedQA}} & \multicolumn{2}{c|}{MedMCQA}    & \multicolumn{1}{c|}{\multirow{2}{*}{sciq}} & ScienceQA      & ScienceQA      & ScienceQA      & \multirow{2}{*}{SimpleQA} \\
                               & \multicolumn{3}{c|}{}                                                 & \multicolumn{1}{c|}{}                          & Single         & Multi          & \multicolumn{1}{c|}{}                      & NAT            & SOC            & LAN            &                           \\ \hline
\diagbox{Method}{Metric}                        & Pre              & Rec              & \multicolumn{1}{c|}{BertScore}      & \multicolumn{1}{c|}{Acc}                       & Acc            & Acc            & \multicolumn{1}{c|}{Acc}                   & Acc            & Acc            & Acc            & Acc                       \\ \hline
GPT-3.5                        & 0.775          & 0.792          & \multicolumn{1}{c|}{0.782}          & \multicolumn{1}{c|}{0.157}                     & 0.512          & 0.507          & \multicolumn{1}{c|}{0.909}                 & 0.842          & 0.884          & 0.724          & 0.220                     \\
GPT-4o                         & 0.759          & 0.798          & \multicolumn{1}{c|}{0.777}          & \multicolumn{1}{c|}{0.175}                     & 0.708          & 0.654          & \multicolumn{1}{c|}{0.966}                 & 0.960          & \textbf{0.995} & 0.862          & 0.282                     \\
CoT \cite{wei2022chain}        & 0.757          & 0.799          & \multicolumn{1}{c|}{0.777}          & \multicolumn{1}{c|}{0.372}                     & 0.538          & 0.497          & \multicolumn{1}{c|}{0.935}                 & 0.824          & 0.876          & 0.768          & 0.203                    \\
ToG \cite{sun2023think}        & 0.741          & 0.800          & \multicolumn{1}{c|}{0.769}          & \multicolumn{1}{c|}{0.210}                     & 0.576          & 0.507          & \multicolumn{1}{c|}{0.933}                 & 0.864          & 0.800          & 0.778          & \textbf{0.536}           \\ 
\rowcolor{gray!8} \textbf{WTS(GPT-3.5)} & \textbf{0.781} & \textbf{0.810} & \multicolumn{1}{c|}{\textbf{0.795}} & \multicolumn{1}{c|}{0.320}                     & 0.622          & 0.557          & \multicolumn{1}{c|}{0.942}                 & 0.890          & 0.964          & 0.810          & 0.264                     \\
\rowcolor{gray!8} \textbf{WTS(GPT-4o)}  & 0.774          & 0.809          & \multicolumn{1}{c|}{0.791}          & \multicolumn{1}{c|}{\textbf{0.397}}            & \textbf{0.781} & \textbf{0.728} & \multicolumn{1}{c|}{\textbf{0.971}}        & \textbf{0.980} & \textbf{0.995} & \textbf{0.906} & 0.306                     \\ \hline
\end{tabular}}

\end{threeparttable}
\vspace{-0.1 in}
\end{table*}

Table \ref{tab:main results} compares the performance of WTS and baselines. The results indicate that WTS (with GPT-4o) outperforms the baselines in 5 of 6 evaluated datasets. The exception occurs on the SimpleQA which is a general domain Q\&A dataset, and in this case, ToG achieves the best performance as it includes a well-established general KG, i.e., Freebase KG, for knowledge retrieval. 

WTS demonstrates a substantial performance improvement over the standard I/O prompt methods (i.e., GPT-3.5 and GPT-4o) on all 5 domain datasets. The highest performance improvement is achieved on the PubMedQA, providing 103.8\% and 126.9\% accuracy gain for GPT-3.5 and GPT-4o, respectively. However, it is noteworthy that in the ChatDoctor5k dataset, GPT-4o underperforms GPT-3.5. This anomaly is attributed to the conservation of GPT-4o, wherein it frequently qualifies its responses as non-professional advice and recommends patients to consult a medical professional.

WTS exhibits notable advantages in comparison with CoT across all evaluated datasets. The most significant performance improvements are 15.6\% and 12.1\% on the single-choice and multiple-choice MedMCQA datasets, respectively. These enhancements are attributable to the incorporation of external domain knowledge. In contrast, CoT relies solely on the intrinsic knowledge of the LLM, which lacks domain-specialized information.

For commonsense reasoning (i.e., SimpleQA), ToG outperforms other methods as it utilizes commonsense knowledge retrieved from Freebase. However, when applied to specialized domains, the performance of ToG becomes inferior because the knowledge in Freebase is coarse and insufficient for domain-specialized questions. 

\begin{figure}[htbp]
  \centering
   \vspace{-0.05 in}
  \begin{minipage}[t]{0.45\linewidth}
    \centering    \includegraphics[width=\linewidth]{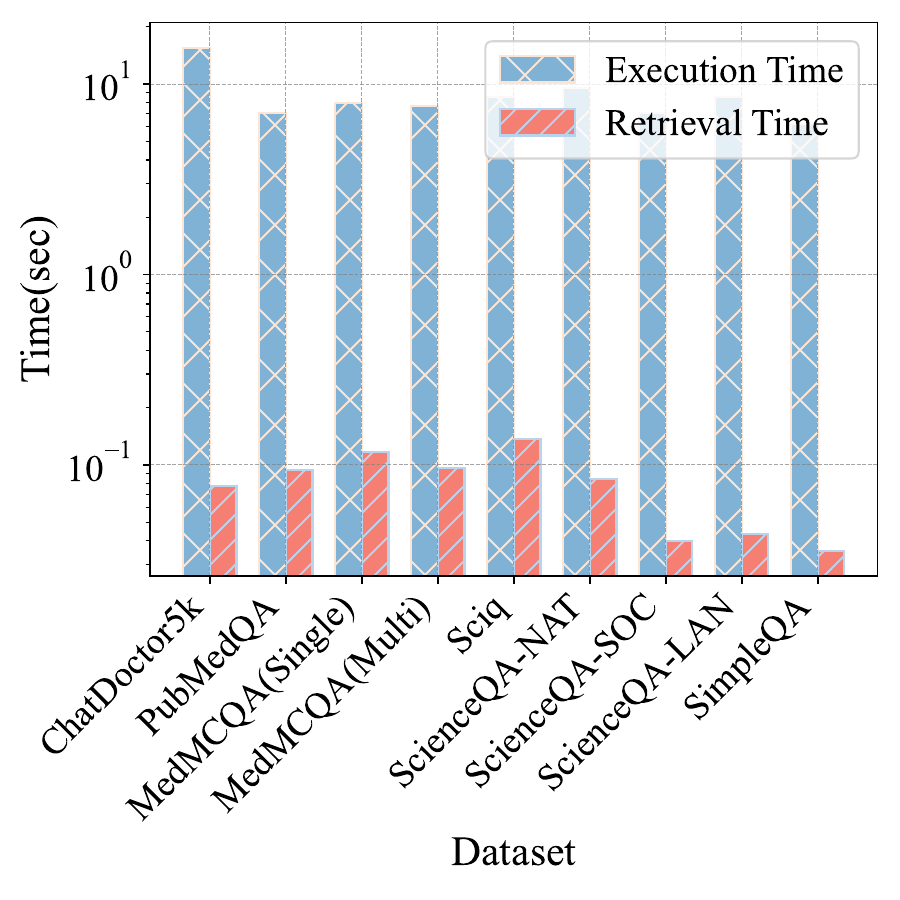}
    \vspace{-0.2 in}
    \caption{Retrieval time and execution time of WTS with GPT-3.5.}
    \label{fig.time_contrast}
  \end{minipage}
  \hspace{0.04\linewidth}
  \begin{minipage}[t]{0.45\linewidth}
    \centering
    \includegraphics[width=\linewidth]{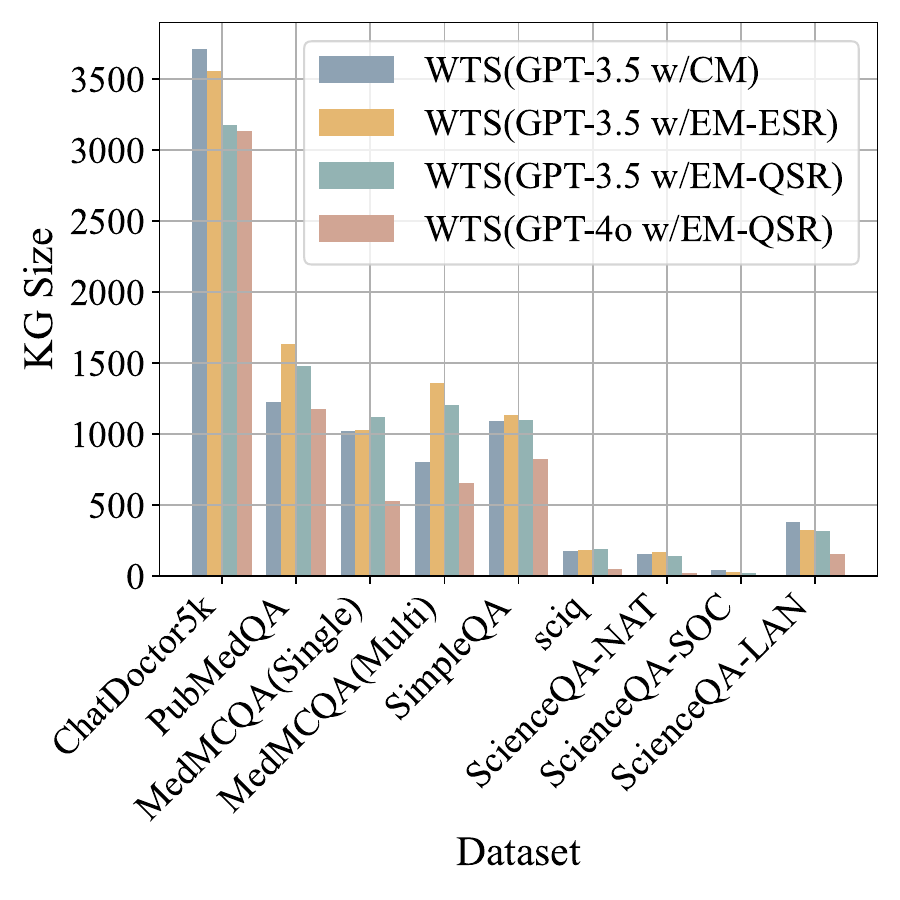}
    \vspace{-0.2 in}
    \caption{DKG Size of WTS with GPT-3.5 and GPT-4o.}
    \label{fig.kg_size}
  \end{minipage}
  \vspace{-0.1in}
\end{figure}

To examine the complexity of the knowledge retrieval process, we calculate the average retrieval time and the execution time of DKG-Augmented LLM (with GPT-3.5) for a single question. The result is depicted in Figure \ref{fig.time_contrast}, which reveals that the execution time is two orders of magnitude greater than the retrieval time, indicating that the retrieval time has a negligible impact on the overall efficiency of WTS.

\begin{figure}[tb]
  \centering
  \subfigbottomskip = 0pt 
  \subfigcapskip = -7pt 
\vspace{-0.1 in}
   \subfigure[GPT-3.5]{
   \includegraphics[width=\linewidth]{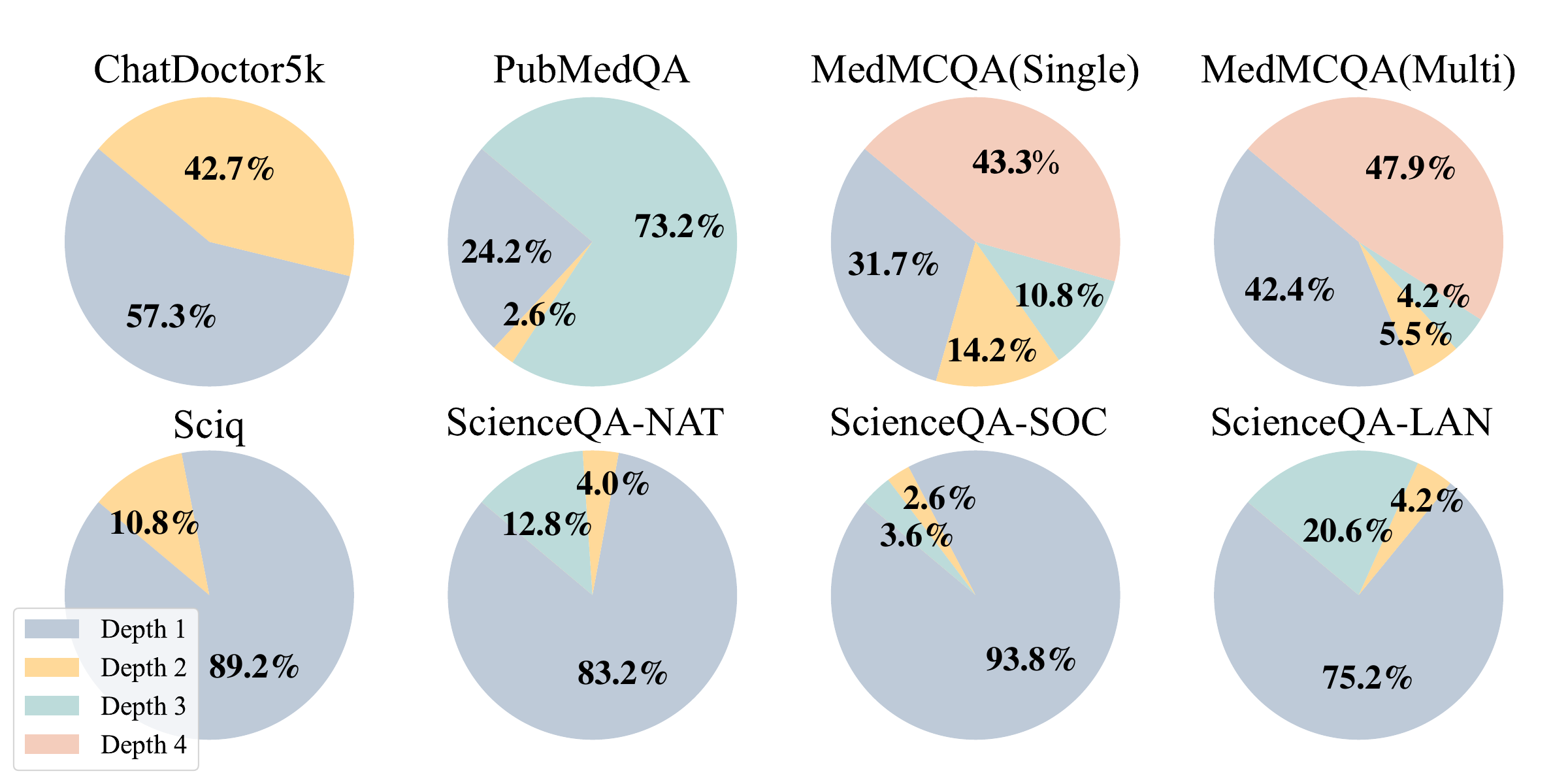}}
   \vspace{-4mm}

   \subfigure[GPT-4o]{
   \includegraphics[width = \linewidth]{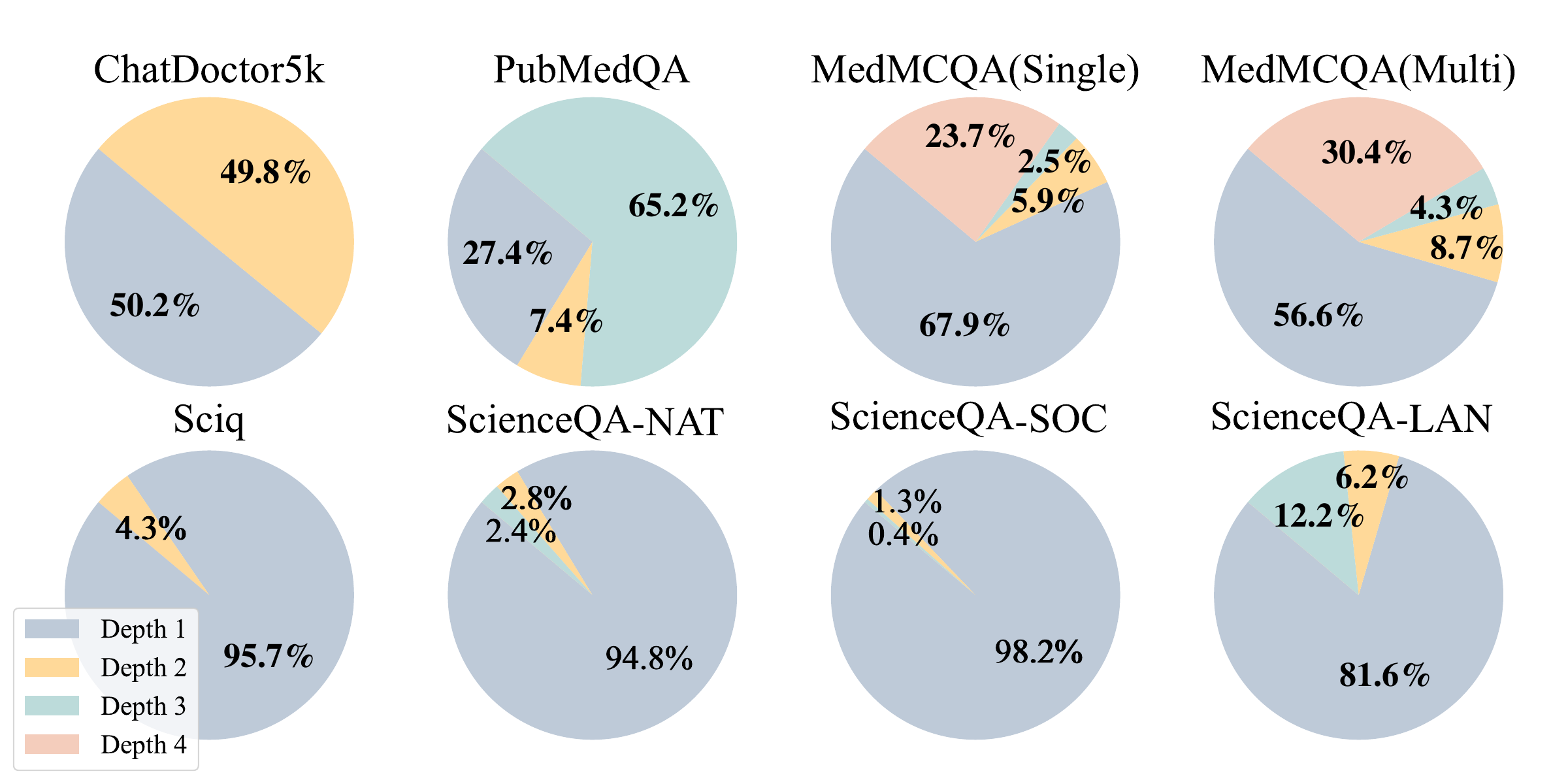}}

\vspace{-0.05 in}
\caption{Percentage of questions across different retrieval depths with WTS (GPT-3.5) and WTS (GPT-4o).}
\vspace{-0.3 in}
  \label{fig.retrival_depth}
\end{figure}

We further investigate the impact of base LLM models on the retrieval process of WTS. Figure  \ref{fig.retrival_depth} illustrates the retrieval depth for questions when WTS is implemented with GPT-3.5 and GPT-4o. The results indicate that a more powerful base model can better leverage its internal knowledge, thereby avoiding deep retrieval of domain knowledge in DKG. Additionally, the enhanced text understanding and summarization abilities of stronger base models facilitate superior knowledge extraction, leading to improved performance.

\subsubsection{Ablation Study}\

\textbf{How does the retrieval mechanism influence the performance of WTS?}
We conduct ablation studies on the retrieval mechanism by executing WTS with 3 alternative retrieval methods. 

\begin{table*}
\caption{Impact of retrieval mechanisms.}
\vspace{-0.1 in}
  \label{tab:retrieval}
  \Description{description}
\resizebox{\textwidth}{!}{
\begin{tabular}{c|cccccc|cc|c|c|c}
\hline
Domain                   & \multicolumn{6}{c|}{Medical}                                                                                                                             & \multicolumn{2}{c|}{Natural}                                & Social         & Linguistic      & General                \\ \hline
\multirow{2}{*}{Dataset} & \multicolumn{3}{c|}{\multirow{2}{*}{ChatDoctor5k}}                    & \multicolumn{1}{c|}{\multirow{2}{*}{PubMedQA}} & \multicolumn{2}{c|}{MedMCQA}    & \multicolumn{1}{c|}{\multirow{2}{*}{sciq}} & ScienceQA      & ScienceQA      & ScienceQA      & \multirow{2}{*}{SimpleQA} \\
                         & \multicolumn{3}{c|}{}                                                 & \multicolumn{1}{c|}{}                          & Single         & Multi          & \multicolumn{1}{c|}{}                      & NAT            & SOC            & LAN            &                           \\ \hline
\diagbox{Method}{Metric}                   & Pre              & Rec              & \multicolumn{1}{c|}{BertScore}      & \multicolumn{1}{c|}{Acc}                       & Acc            & Acc            & \multicolumn{1}{c|}{Acc}                   & Acc            & Acc            & Acc            & Acc                       \\ \hline
GPT-3.5                  & 0.775          & 0.792          & \multicolumn{1}{c|}{0.783}          & \multicolumn{1}{c|}{0.157}                     & 0.512          & 0.507          & \multicolumn{1}{c|}{0.909}                 & 0.842          & 0.884          & 0.724          & 0.220                     \\
WTS(EM)                  & 0.775          & 0.809          & \multicolumn{1}{c|}{0.791}          & \multicolumn{1}{c|}{0.281}                     & 0.576          & 0.548          & \multicolumn{1}{c|}{0.915}                 & 0.874          & 0.933          & 0.786          & 0.256                     \\
WTS(EM-ESR)              & 0.778          & 0.808          & \multicolumn{1}{c|}{0.792}          & \multicolumn{1}{c|}{0.293}                     & 0.604          & \textbf{0.563} & \multicolumn{1}{c|}{0.917}                 & 0.876          & 0.960          & 0.808          & \textbf{0.264}            \\
WTS(EM-QSR)              & \textbf{0.781} & \textbf{0.810} & \multicolumn{1}{c|}{\textbf{0.795}} & \multicolumn{1}{c|}{\textbf{0.320}}            & \textbf{0.622} & 0.557          & \multicolumn{1}{c|}{\textbf{0.942}}        & \textbf{0.890} & \textbf{0.964} & \textbf{0.810} & \textbf{0.264}            \\ \hline
\end{tabular}}
\vspace{-0.1 in}
\end{table*}

(1) \textbf{Exact Match (EM).} This method retrieves knowledge triples that have the target entity as their subject or object entity.

(2) \textbf{Exact Match with Entity Similarity Retrieval (EM-ESR).} This method first uses the exact match method for knowledge triple retrieval. The retrieved triple will then be refined by evaluating the similarity between the embedding of extracted topic entities and knowledge triples retrieved by exact match. 

(3) \textbf{Exact Match with Question Similarity Retrieval (EM-QSR).} This method follows a similar procedure as EM-ESR. The key difference is that EM-QSR calculates the similarity between the embedding of questions and knowledge triples.

Table \ref{tab:retrieval} presents the performance of WTS with different retrieval mechanisms configured in DKG-Augmented LLM. We see that EM-QSR exhibits the highest performances on most datasets as it fully utilizes the entities and other semantic information in question to identify most relevant knowledge triples. EM-ESR ranks second because it captures less semantic information compared to EM-QSR. The inferior performance of EM indicates that relying solely on exact matching and neglecting the semantic information of the question fails to obtain high-quality knowledge.

\begin{figure}[htbp]
  \centering
  \subfigbottomskip = 0.5pt 
  \subfigcapskip = -3 pt 
  \subfigure[ChatDoctor5k]{
    \includegraphics[width = 0.45\linewidth]{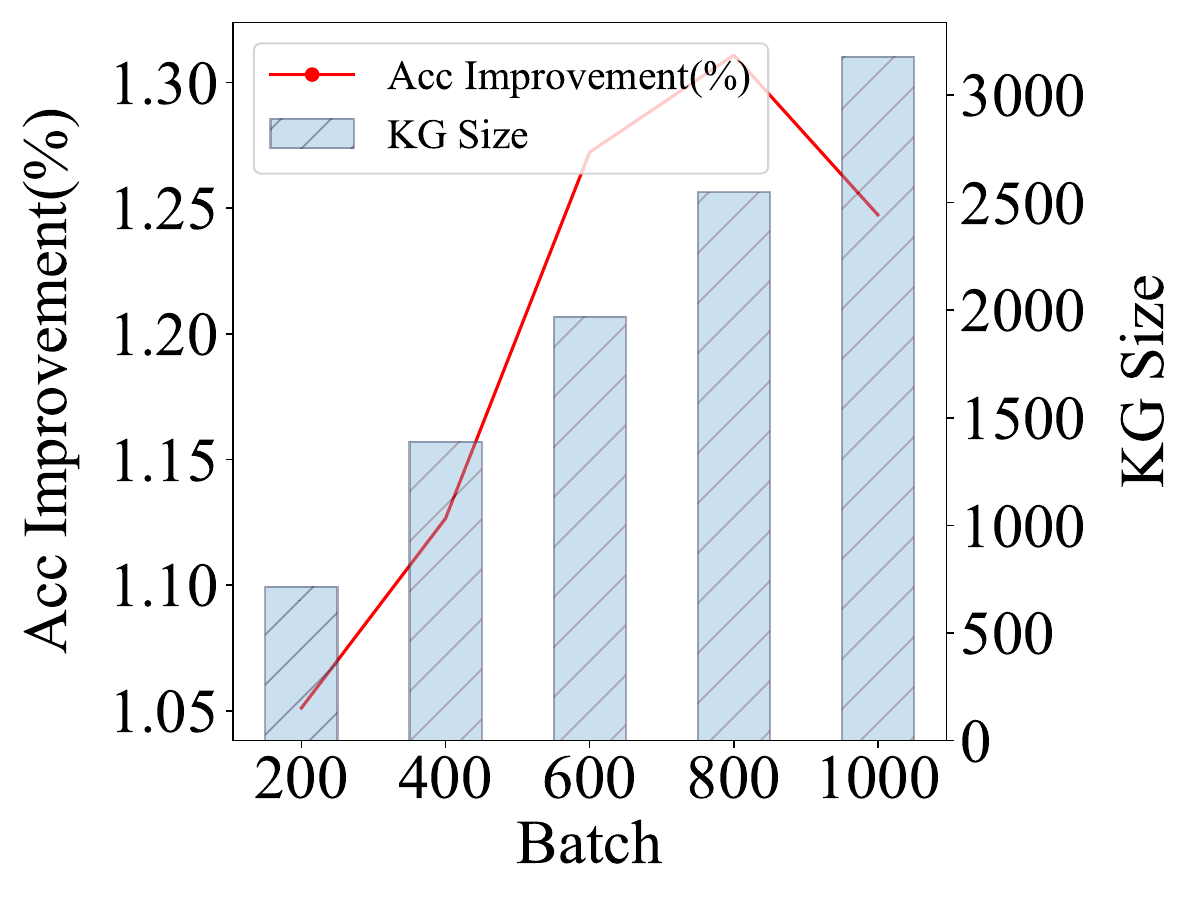}}
    \hspace{0.05 in}
   \subfigure[PubMedQA]{
   \includegraphics[width = 0.45\linewidth]{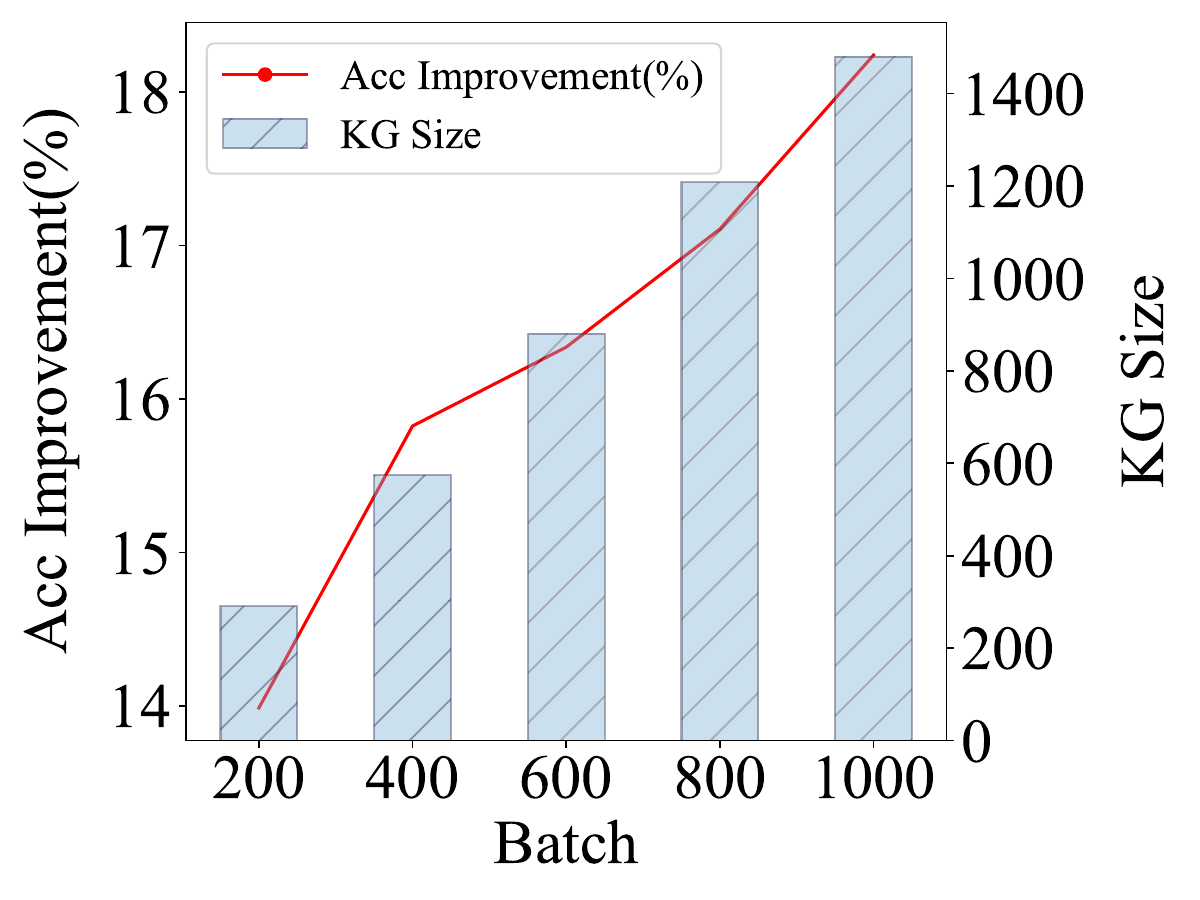}
}
\vspace{-1mm}

\subfigure[MedMCQA(Single)]{
    \includegraphics[width = 0.45\linewidth]{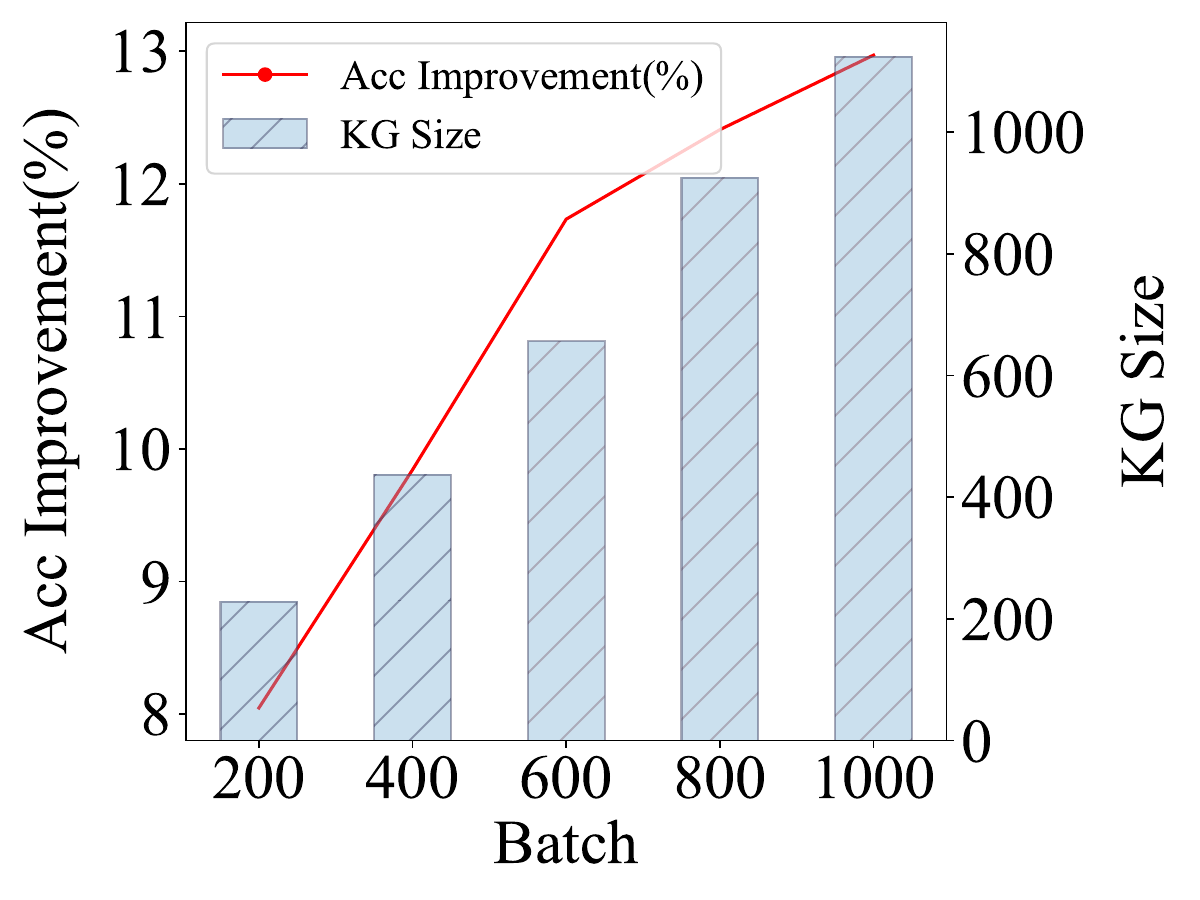}}
    \hspace{0.05 in}
   \subfigure[MedMCQA(Multi)]{
   \includegraphics[width = 0.45\linewidth]{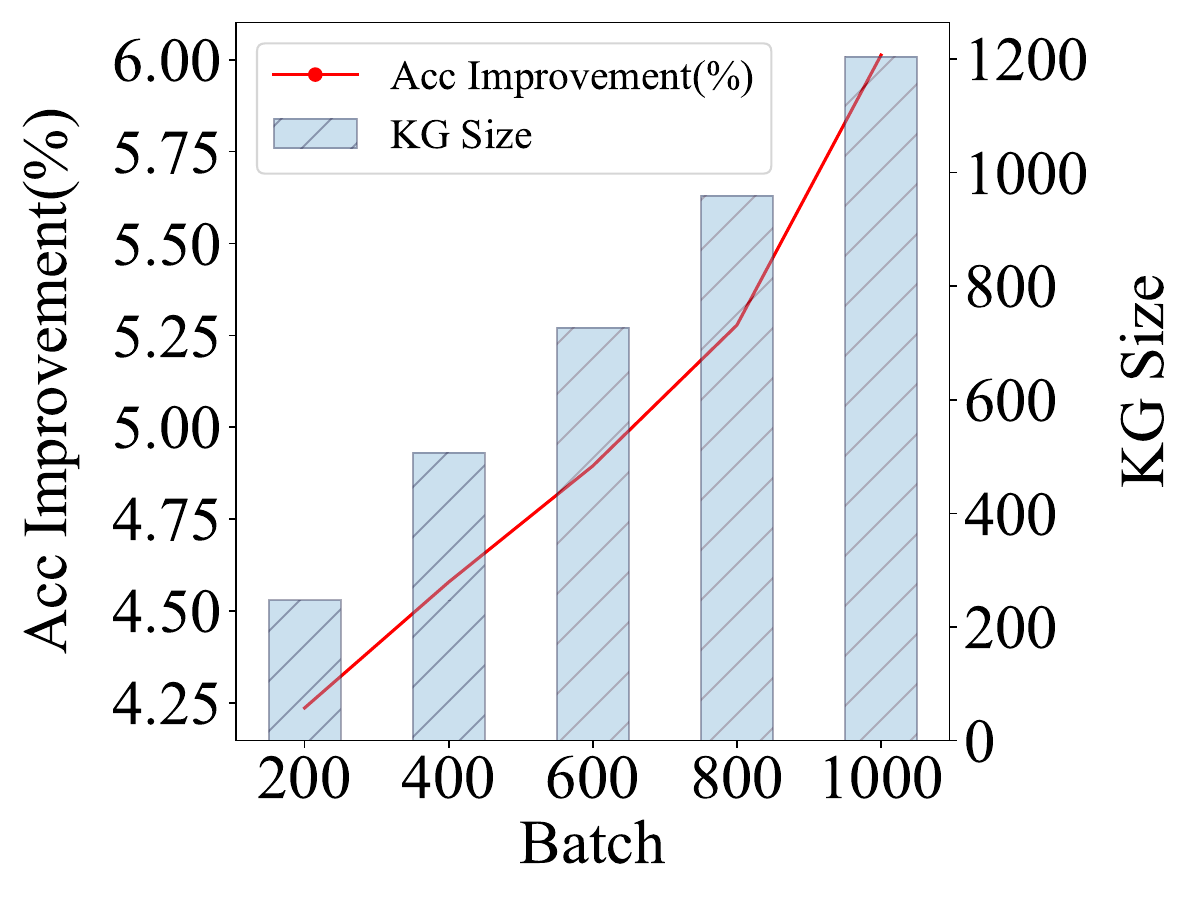}
}

\vspace{-1mm}
\subfigure[Sciq]{
    \includegraphics[width = 0.45\linewidth]{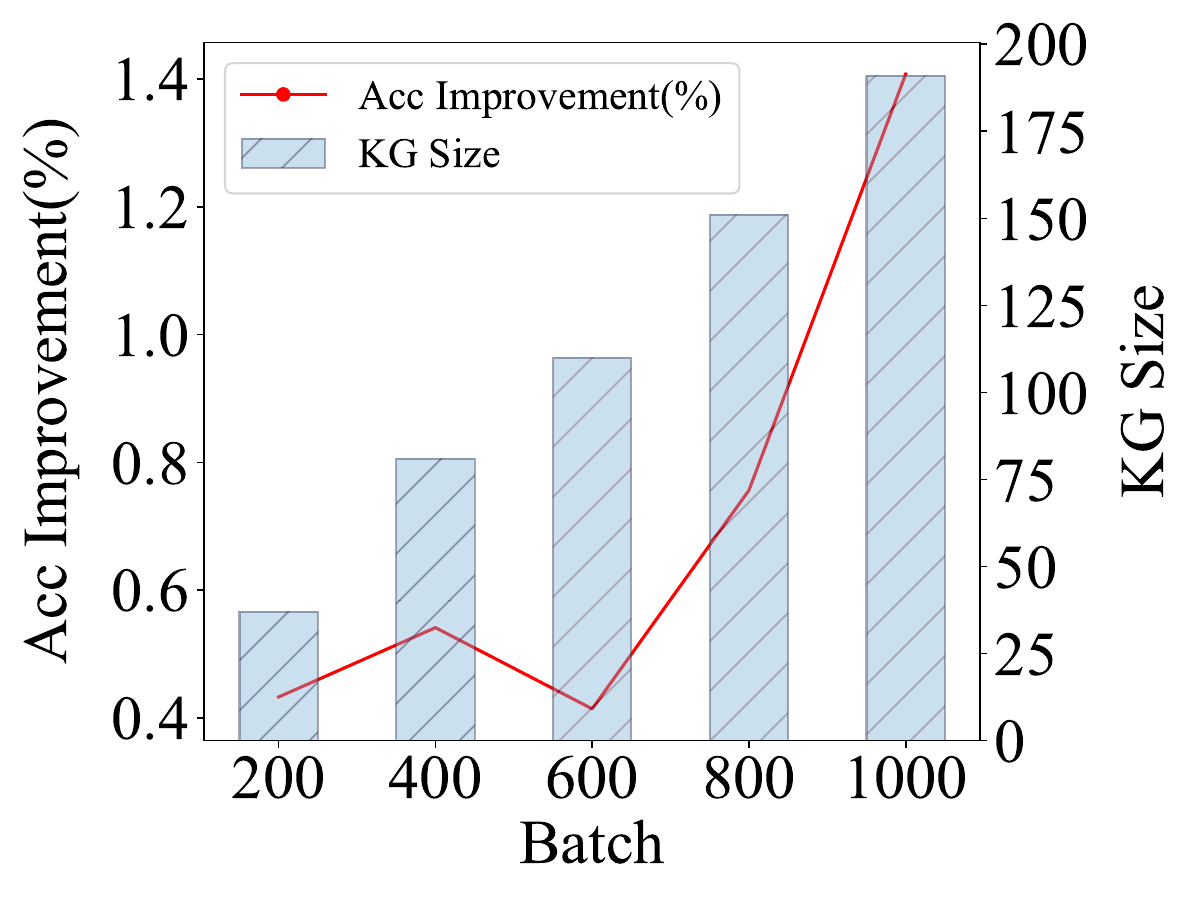}}
    \hspace{0.05 in}
    \subfigure[ScienceQA-NAT]{
    \includegraphics[width = 0.45\linewidth]{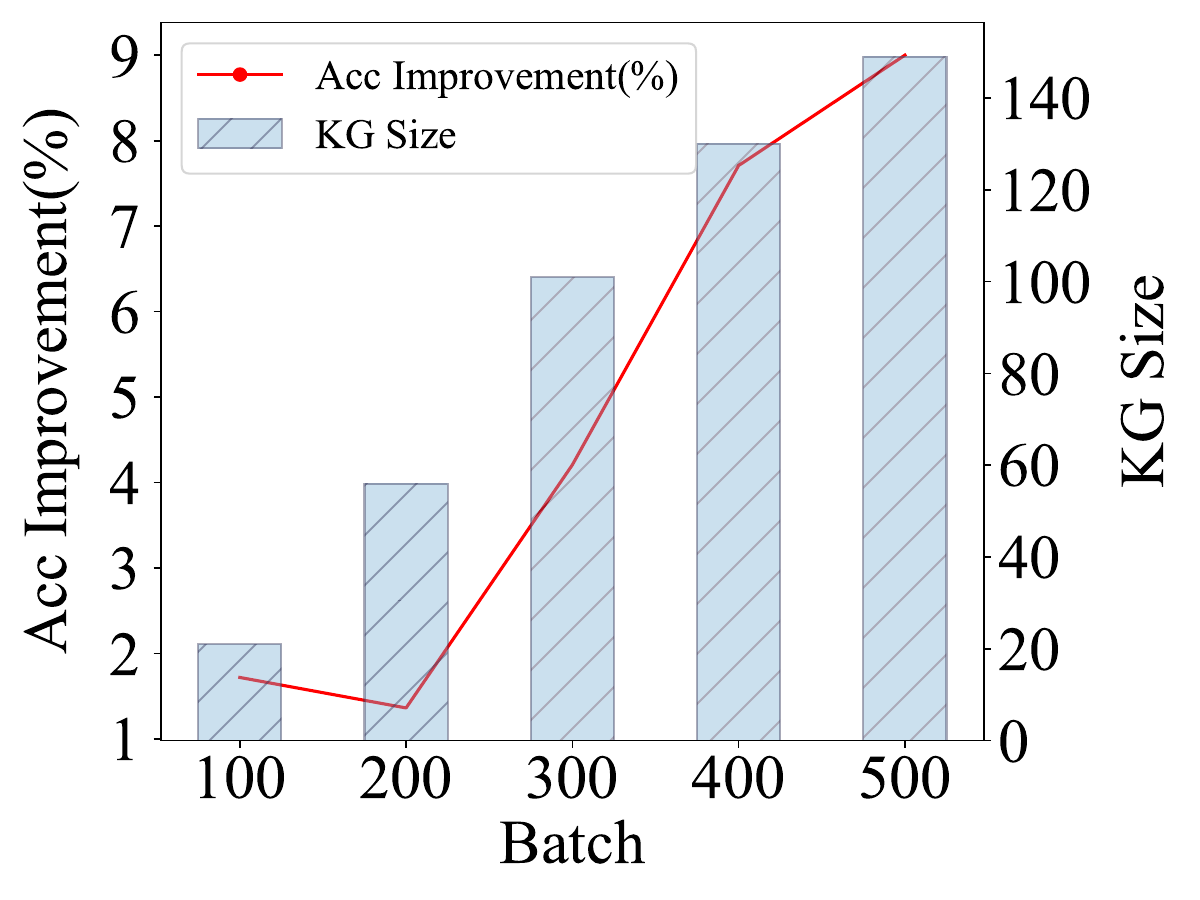}}
\vspace{-1mm}
    \subfigure[ScienceQA-SOC]{
   \includegraphics[width = 0.45\linewidth]{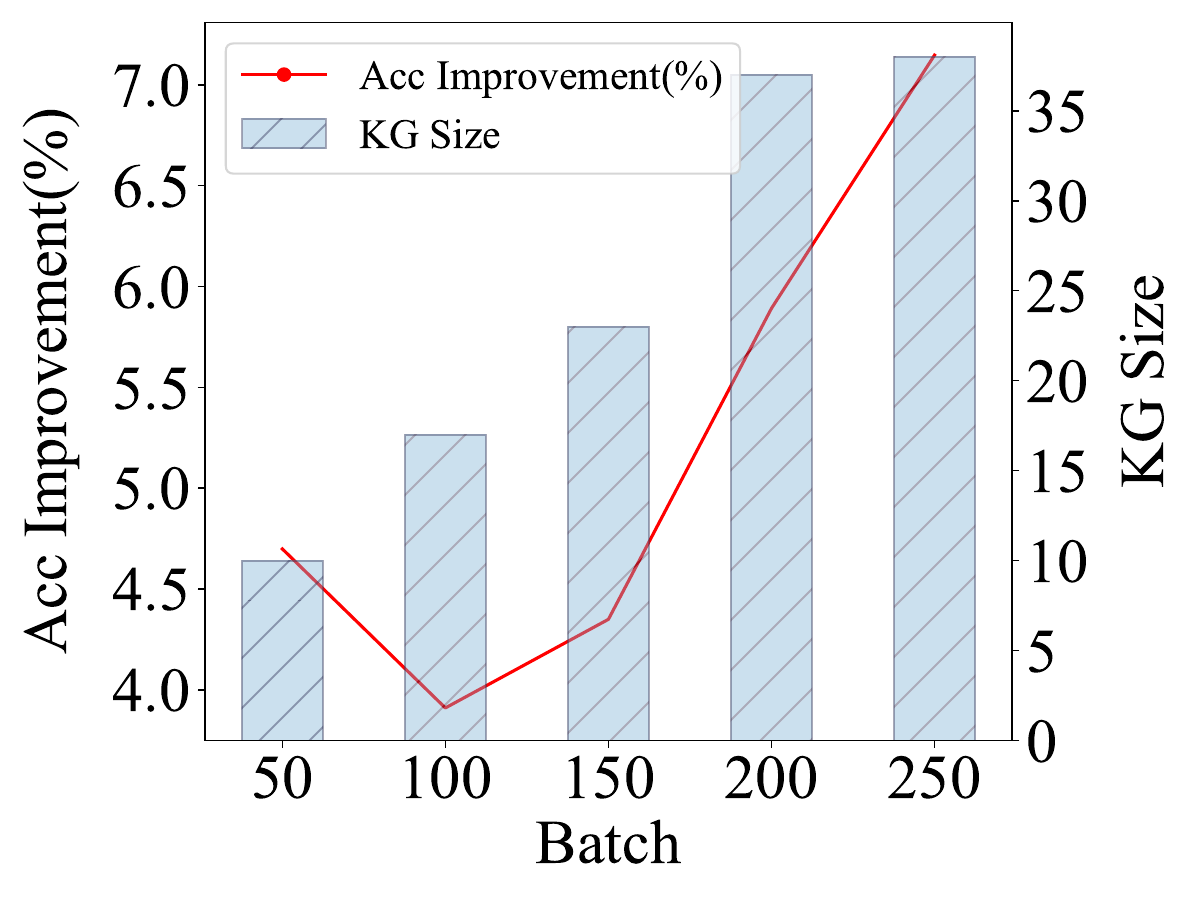}}
    \hspace{0.05 in}
   \subfigure[ScienceQA-LAN]{
   \includegraphics[width = 0.45\linewidth]{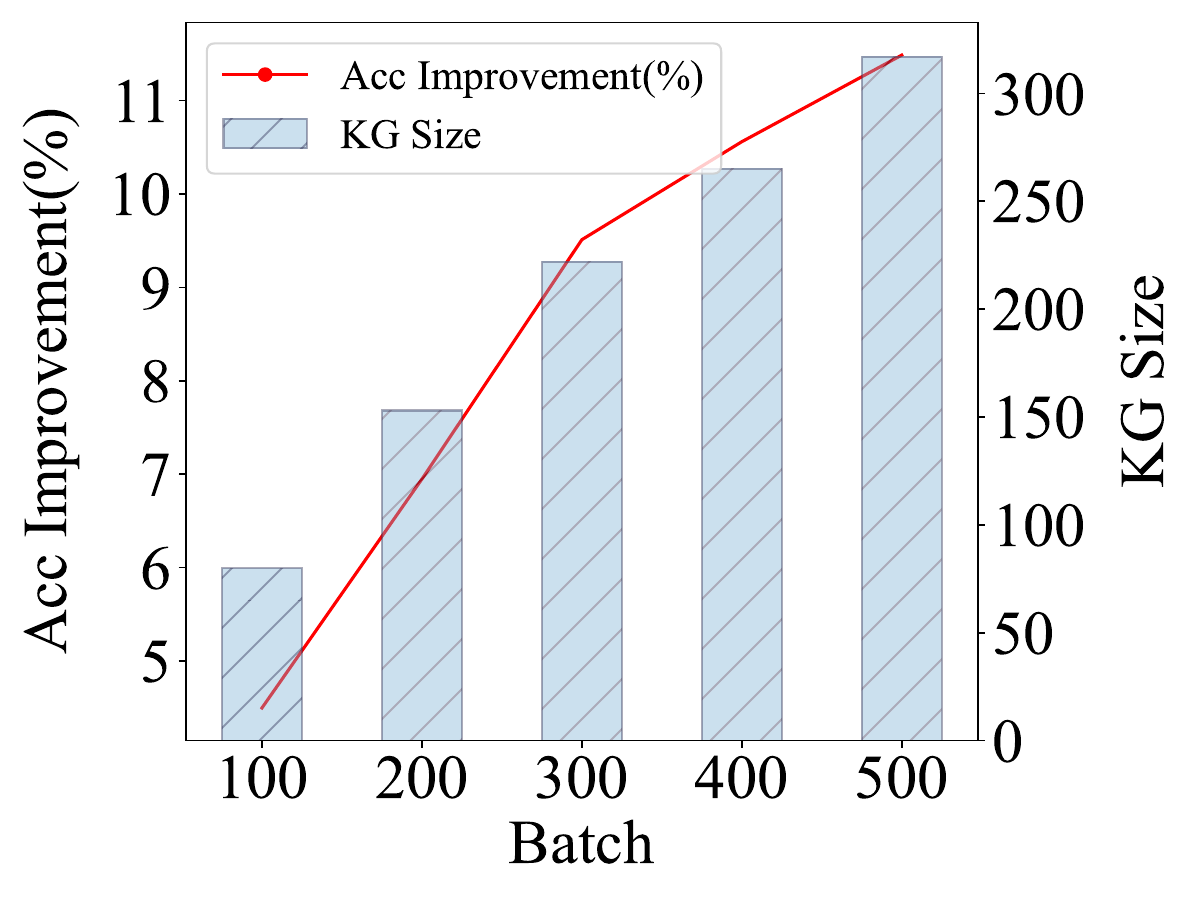}
}
\vspace{-0.05 in}
  \caption{Accuracy improvements and KG size in medical domain.}
  \vspace{-0.25 in}
  \label{fig.dif_line}
\end{figure}

\textbf{How does DKG evolution affect the WTS performance?}
Next, we show how DKG evolution benefits the reasoning performance of specialized LLM. Figure \ref{fig.dif_line} shows the accuracy improvement of WTS over GPT-3.5 as WTS proceeds. It is evident that as the number of processed Q\&A samples increases, the size of DKG increases accordingly. 
It can be observed that the performance improvement experiences fluctuations during the formation process of WTS. This is attributed to the uncertain knowledge overlap between previously answered questions and questions yet to be answered. Despite these fluctuations, the overall trend indicates an enhancement in generation quality. Figure \ref{fig.kg_size} illustrates the sizes of DKGs constructed by WTS with different base models and retrieval mechanisms. Disregarding the minor randomness introduced by the ``temperature parameter'' of LLMs, we see that employing advanced models and retrieval mechanisms results in the construction of compact DKGs. This is because an advanced base model possesses more comprehensive inherent knowledge, and an advanced retrieval model can acquire relevant knowledge triples from DKG with greater efficiency.
\begin{table}[!htbp]
    \centering
    \vspace{-0.08 in}
        \caption{Accuracy of WTS with different retrieval depth.}
        \vspace{-0.1 in}
    \resizebox{0.85\linewidth}{!}{
   \begin{tabular}{c|cccc}
\hline
Dataset         & D1   & D2            & D3            & D4           \\ \hline
Chatdoctor5k    & 0.792 & \textbf{0.795} & 0.794          & 0.794             \\
PubMedQA        & 0.211 & 0.256          & \textbf{0.320} & 0.275     \\
MedMCQA(Single)    & 0.358          & 0.457          & \textbf{0.622} & 0.550 \\
MedMCQA(Multi)     & 0.360          & 0.493          & \textbf{0.557} & 0.527 \\ \hline
\end{tabular}}
\vspace{-0.1 in}
    \label{tab:dif_depth}
\end{table}

\textbf{How does the retrieval depth affect the WTS performance?}
To investigate the impact retrieval Depth $D$, we conducted experiments of WTS with various depths. For ChatDoctor5k and PubMedQA, the maximum retrieval depth ranges from 1 to 4, while for MedMCQA, it is set from 2 to 5.
Table \ref{tab:dif_depth} illustrates the performance on ChatDoctor5k, PubMedQA, and MedMCQA datasets with varying retrieval depths. As shown in the results, the performance of WTS improves with increased search depth, suggesting that further enhancement could be achieved with deeper exploration in DKG. However, this improvement decreases as the retrieval depth becomes excessive. This is primarily because deeper retrieval introduces more irrelevant information, which disrupts the generation of accurate answers and increases computational overhead. The execution time of the different maximal retrieval depths on these datasets is shown in Appendix \ref{result-analysis} Table \ref{tab:dif_depth_time}.

\section{Conclusion}

In this paper, we introduced the LLM$\circlearrowright$KG paradigm, which achieves bidirectional enhancement between LLMs and knowledge graphs. We proposed the flexible plug-and-play Way-to-Specialist (WTS) framework, combining specialized LLMs with evolving domain knowledge graphs to address knowledge-intensive tasks in specific domains. Through its two tightly integrated components—the DKG-Augmented LLM and the LLM-Assisted DKG Evolution—WTS demonstrates superior performance over existing prompting-based methods in domain-specific tasks, all while avoiding additional training costs. Our experimental results demonstrate that WTS surpasses the previous state-of-the-art in four specialized domains, achieving a maximum performance improvement of 11.3\%. 
In future work, we will focus on developing more efficient and high-quality approaches for dynamic knowledge graph completion.

\bibliographystyle{ACM-Reference-Format}
\bibliography{sample-base}
\newpage

\appendix

\section{ALGORITHM FOR WTS}
\label{algorithm}

We present a detailed overview of the algorithmic progress of WTS, as shown on Algorithm \ref{alg:one}. 

\RestyleAlgo{ruled}
\SetKwComment{Comment}{/* }{ */}

\begin{algorithm}[hbt!]
\caption{The workflow of WTS}\label{alg:one}
\KwData{\textcolor{gray}{\texttt{question} $q$, \texttt{real\_answer} $\alpha^*_{q}$, \texttt{domain knowledge graph} $\mathcal{G}$, \texttt{max\_depth} $D$, \texttt{max\_width} $W$, \texttt{max\_hop} $H$ ($H$<$D$) }}
\KwResult{\textcolor{gray}{LLM answer $\alpha_{q}$}}
$\mathcal{E}_q \gets \texttt{LLM\_Entity\_Extract}(q)$\;
\texttt{depth} $ d \gets 1$\;
$\mathcal{\bar{T}}_q \gets \{\}$\;
$\mathcal{E}^{(1)}_q \gets \mathcal{E}_q$\;
\While{$d\leq D$}{
  $\mathcal{T}_{q}^{(d)} \gets \texttt{Retrieve\_KG}(\mathcal{E}^{(d)},\mathcal{G})$ \textcolor{gray}{\# Retrieve entities in DKG}\;  
  $\mathcal{\tilde{T}}_{q}^{(d)} \gets \texttt{LLM\_Score\_and\_Prune}(\mathcal{T}_{q}^{(d)}, q, W)$ \textcolor{gray}{\# Use LLM to score and prune the triples}\;   
  $\mathcal{\bar{T}}_q \gets \bigcup^{d}_{i=1} \mathcal{\tilde{T}}_{q}^{(i)}$ \textcolor{gray}{\# Update the retrieved knowledge triples}\; 
  $\mathcal{E}^{(d+1)}_q = \{ e | e \in \mathcal{\tilde{T}}^{(d)}_q, e \notin \bigcup^{d}_{i=1} \mathcal{E}^{(i)}_q\}$\textcolor{gray}{\# Update the retrieved entities}\;
  $\texttt{confidence},\alpha_{q}^{(d)} \gets \texttt{LLM\_Generation}(q,\mathcal{\bar{T}}_q)$ \textcolor{gray}{\# Prompt LLM to generate answer}\;
  {\eIf{\texttt{confidence is positive and} $d \leq H$}{
  $\alpha_{q} \gets \alpha_{q}^{(d)}$ \textcolor{gray}{\# Get the final answer of LLM}\; 
  break\;
  }
    {\eIf{\texttt{confidence is positive and} $d > H$}{
    $\alpha_{q} \gets \alpha_{q}^{(d)}$\;
      $\mathcal{T}^{+}_q \gets \texttt{LLM\_Create\_Triple}(q,\alpha^*_{q},\mathcal{\bar{T}}_q)$ \textcolor{gray}{\# Prompt LLM to generate KG triples}\;
      $\mathcal{G} \gets \texttt{Update\_KG}(\mathcal{G},\mathcal{T}^{+}_q)$ \textcolor{gray}{\# Update KG with generated triples}\;
      break;
    }
    {\If{\texttt{confidence is negative and} $d == D$}
    {
    $\alpha_{q} \gets \alpha_{q}^{(d)}$\;
      $\mathcal{T}^{+}_q \gets \texttt{LLM\_Create\_Triple}(q,\alpha^*_{q},\mathcal{\bar{T}}_q )$ \;
      $\mathcal{G} \gets \texttt{Update\_KG}(\mathcal{G},\mathcal{T}^{+}_q)$\;
    }
    }
    }
}}
\end{algorithm}

\section{Experiment Details}

\subsection{Dataset}
\label{Dataset}

The statistics of the datasets used in this paper are shown in Table \ref{tab:dataset}. To avoid the influence of context provided by the dataset on the evaluation of methods, no extra context information except the retrieved results is offered as the support information during the experiments. The detailed descriptions of their usage in this paper are as follows:

\begin{table*}[!htpb]
\caption{Statistics of datasets used in experiments.}
    \label{tab:dataset}
    \Description{description}
\begin{tabular}{c|ccccl}
\hline
Dataset      & Answer Format       & Train   & Validation & Test  & Details                                                                                                           \\ \hline
ChatDoctor5k & Text Generation     & 5,452   & -          & -     & -                                                                                                                 \\ \hline
PubMedQA     & True/False Question & 273,518 & -          & -     & \begin{tabular}[c]{@{}l@{}}Artificial Generated: 211,269\\ Expert Labeled: 1,000\\ Unlabeled: 61,249\end{tabular}              \\ \hline
MedMCQA      & Multiple Choice Question     & 182,822 & 6,150      & 4,183 & \begin{tabular}[c]{@{}l@{}}Single Choice Question: 127,715\\ Multiple Choice Question: 65,400\end{tabular}        \\ \hline
Sciq         & Multiple Choice Question     & 11,679  & 1,000      & 1,000 & -                                                                                                                 \\ \hline
ScienceQA    & Multiple Choice Question     & 12,726  & 4,241      & 4,241 & \begin{tabular}[c]{@{}l@{}}Social Science: 4,350\\ Natural Science: 11,487\\ Language Science: 5,371\end{tabular} \\ \hline
SimpleQA     & Multiple Choice Question     & 14,894  & -          & 1,000 & -                                                                                                                 \\ \hline
\end{tabular}
\end{table*}

\textbf{ChatDoctor5k}\quad ChatDoctor5k \cite{li2023chatdoctor} is a generated conversation between patients and physicians from ChatGPT grounded on a disease database. Each input describes the patient's symptoms possibly with past medical history or etiological factor attached. And the desired output needs to give professional suggestions covering the diagnosis, symptoms, recommended treatments, medical tests, and so on. We sampled 1000 dialogues as the test set to test the long text generation capability of WTS.

\textbf{PubMedQA}\quad PubMedQA \cite{jin2019pubmedqa} contains tests requiring to answer biomedical research questions with yes, no or maybe given the context provided by PubMed abstracts. To fully display the performance of knowledge augmentation, we adopt no given context and reasoning-free setting and use the labeled 1000 Q\&A pairs in the experiment.

\textbf{MedMCQA}\quad MedMCQA \cite{pmlr-v174-pal22a} is a large-scale, Multiple-Choice Question Answering (MCQA) dataset designed to address real-world medical entrance exam questions. We sample 1000 single-choice questions and 1000 multiple-choice questions from the “dev” subset of the dataset with random 2 to 4 choices as the test set.

\textbf{SciQ}\quad SciQ \cite{Welbl2017CrowdsourcingMC} is a multiple-choice format science exam dataset about physics, chemistry and biology. Each question has 4 answer options. We use the "test" subset of SciQ containing 1000 questions and provide no additional paragraph with supporting evidence to evaluate our methods.

\textbf{ScienceQA}\quad ScienceQA \cite{saikh2022scienceqa} is a large-scale science question-answering dataset containing various knowledge domains. In our experiment, we sample the 500 natural science questions, 500 language science questions, and 225 social science questions in the ``test'' subset of ScienceQA with no image and no textual context.

\textbf{Simple Questions}\quad Simple Questions \cite{bordes2015large} is a large-scale Q\&A dataset with 100k Q\&A pairs constructed based on Freebase knowledge base. Each pair in Simple Questions has a corresponding fact from Freebase and can be rephrased as the form of (subject, relationship, ? ). We sample 1000 pairs as the test set.

\subsection{Metrics}
\label{Metrics}

For the dataset ChatDoctor5k, we evaluate the text generation task with BERTScore \cite{zhang2019bertscore}, which computes a similarity score for each token in the candidate sentence with each token in the reference sentence. The complete score matches each token in $\mathbf{x}$ to a token in $\hat{\mathbf{x}}$ to compute recall, and each token in $\hat{\mathbf{x}}$ to a token in $\mathbf{x}$  to compute precision. We use greedy matching to maximize the matching similarity score, where each token is matched to the most similar token in the other sentence. We combine precision and recall to compute an F1 measure. For a reference $\mathbf{x}$ and candidate $\hat{\mathbf{x}}$, the recall, precision, and F1 scores are:

\begin{equation}
R_\texttt{BERT} = \frac{1}{|\mathbf{x}|} \sum_{x_i \in \mathbf{x}} \max_{\hat{x}_j \in \hat{\mathbf{x}}} x_i^\top \hat{x}_j,~~ P_\texttt{BERT} = \frac{1}{|\hat{\mathbf{x}}|} \sum_{\hat{x}_j \in \hat{\mathbf{x}}} \max_{x_i \in \mathbf{x}} \hat{x}_i^\top x_j
\end{equation}


\begin{equation}
F_\texttt{BERT} = 2 \times \frac{P_\texttt{BERT} \cdot R_\texttt{BERT}}{P_\texttt{BERT} + R_\texttt{BERT}}
\end{equation}

For the other datasets, we use Accuracy as the metric to evaluate the correctness of the results. Accuracy is defined as the ratio of correctly predicted instances to the total number of instances, 
particularly suitable for multiple-choice questions where the primary concern is the correctness of the predicted labels.

\subsection{Evaluation}
\label{Evaluation}

To facilitate result evaluation, WTS employs prompts to constrain its output format to JSON. To prevent irrelevant generations from interfering with answer evaluation, the LLMs are instructed to provide the label of their answers in multiple-choice questions. Consequently, the evaluation is based solely on whether the label corresponds to the correct answer. This constraint is also applied to the GPT-3.5-turbo and GPT-4o baselines for the sake of fairness. For the ToG and CoT baselines, to avoid imposing format restrictions that may affect their performance, we use a specific signal by adding '(' before the answer and ')' after it. This method ensures the precise extraction of answers and upholds consistency throughout the evaluation process.

\section{Supplementary Results}
\label{result-analysis}

\begin{table}[]
\center
\caption{Total average execution time (sec) and maximal retrieval depth of WTS in each dataset.}
\resizebox{\linewidth}{!}{
\begin{tabular}{cccc}
\hline
Method          & WTS(GPT-3.5)/sec & WTS(GPT-4o)/sec & $D$ \\ \hline
ChatDoctor5k    & 21.26                 & 37.37                & 2         \\
PubMedQA        & 21.93                 & 33.95                & 3         \\
MedMCQA(Single) & 30.38                 & 23.15                & 4         \\
MedMCQA(Multi)  & 17.17                 & 32.50                & 4         \\
sciq            & 9.28                  & 19.17                & 2         \\
ScienceQA(SOC)  & 8.48                  & 21.91                & 3         \\
ScienceQA(NAT)  & 11.84                 & 17.87                & 3         \\
ScienceQA(LAN)  & 11.63                 & 21.35                & 3         \\
SimpleQA        & 14.77                 & 31.31                & 3         \\\hline
\end{tabular}
}

\label{tab:execution time}
\end{table}

Table \ref{tab:execution time} shows the average execution time of WTS for a question, where $D$ represents the maximal retrieval depth. It is apparent that WTS with GPT-4o API incurs a higher time cost compared to GPT-3.5. 

\begin{table*}[h]
\caption{Performance of Apprenticeship and Mastership phase on ChatDoctor5k.}
    \label{tab:unsupervised}
    \Description{description}
\begin{tabular}{c|ccc|ccc}
\hline
Phase                    & \multicolumn{3}{c|}{Apprenticeship}              & \multicolumn{3}{c}{Mastership}                   \\ \hline
\resizebox{0.12\textwidth}{!}{\diagbox{Method}{Metric}} & Pre            & Rec            & BertScore      & Pre            & Rec            & BertScore      \\ \hline
GPT-3.5                  & 0.774          & 0.792          & 0.782          & 0.776          & 0.792          & 0.783          \\
CoT                      & 0.757          & 0.799          & 0.777          & 0.758          & 0.797          & 0.776          \\
ToG                      & 0.741          & 0.800          & 0.768          & 0.741          & 0.800          & 0.769          \\
GPT-4o                   & 0.759          & 0.798          & 0.777          & 0.760          & 0.796          & 0.777          \\
WTS(GPT-3.5)             & \textbf{0.781} & \textbf{0.810} & \textbf{0.794} & \textbf{0.778} & \textbf{0.809} & \textbf{0.792} \\ \hline
\end{tabular}
\end{table*}

Table \ref{tab:unsupervised} shows the performance of the Mastership phase on the ChatDoctor5k dataset. During the Apprenticeship phase, we sampled 800 data to conduct medical Q\&A and consistently evolve the DKG. In the Mastership phase, we tested the performance of WTS with another 200 non-overlapping samples. The results show that WTS achieves state-of-the-art performance compared to the four baseline models.

The execution time (sec) of the different maximal retrieval depths on three medical domain datasets, ChatDoctor5k, PubMedQA, and MedMCQA, is shown in Table \ref{tab:dif_depth_time}. For ChatDoctor5k and PubMedQA, the maximum retrieval depth ranges from 1 to 4, while for MedMCQA, it is set from 2 to 5. The results indicate that as the maximum retrieval depth increases, execution time correspondingly rises. 

\begin{table}[htbp]
\caption{Execution time of different retrieval depth on medical domain datasets.}
    \label{tab:dif_depth_time}
    \Description{description}
\resizebox{0.85\linewidth}{!}{
\begin{tabular}{c|cccc}
\hline
Dataset         & D1     & D2     & D3     & D4       \\ \hline
Chatdoctor5k    & \textbf{15.27} & 21.26 & 21.62 & 23.48      \\
PubMedQA        & \textbf{9.61}  & 12.64 & 21.93 & 18.14      \\
MedMCQA(Single)     & \textbf{15.44} & 16.80 & 30.38 & 37.14 \\
MedMCQA(Multi)      & \textbf{15.50} & 18.77 & 17.17 & 37.57 \\ \hline
\end{tabular}}
\end{table}

\begin{figure}[ht]
\includegraphics[width=0.9\linewidth]{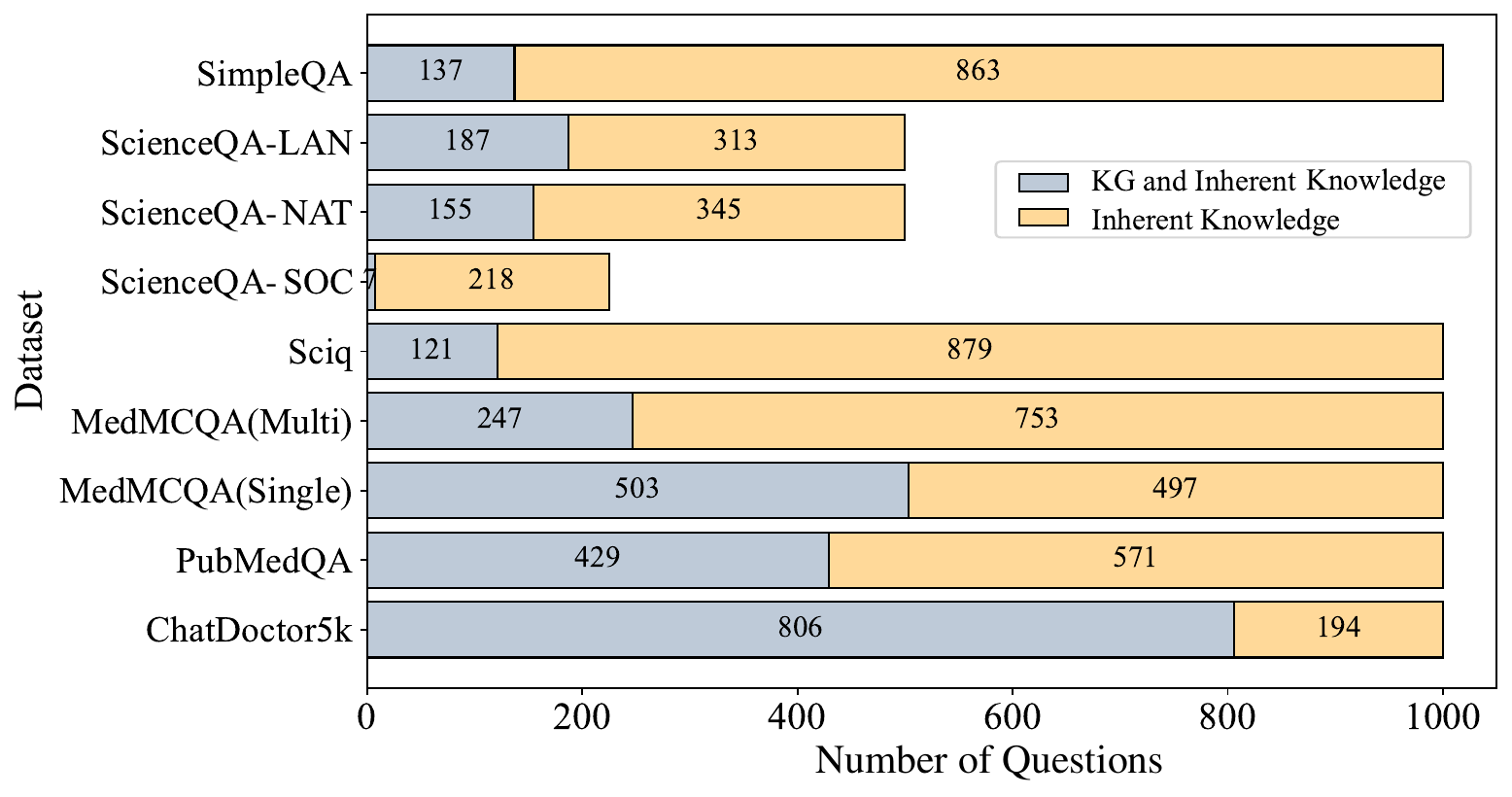}
  \caption{Proportions of WTS's evidence of answers.}
  \Description{The statistic data of WTS's evidence of answers on all the datasets.}
  \label{fig:knowledge_used}
\end{figure}

We analyzed the answers across six datasets to investigate the evidence for WTS in generating answers, as illustrated in Figure \ref{fig:knowledge_used}. The results indicate that in the general domain and certain science domains, e.g., social science, which includes a considerable amount of commonsense questions, a significant proportion of the answers rely exclusively on the intrinsic knowledge embedded within LLM’s parameters. However, in specialized domains such as the medical domain and linguistics, accurate answer generation requires specific domain knowledge. Consequently, a substantial proportion of these queries rely on a combination of knowledge from both retrieved triples and the LLM's inherent knowledge for answer generation.

Table \ref{tab:KG triple count} shows the size of the knowledge graph constructed in the process of WTS. The results show that advanced models with more inherent knowledge lead to the construction of more concise DKGs. Furthermore, the advanced retrieval mechanisms also enhance the conciseness of DKGs by optimizing the WTS's capability to utilize knowledge efficiently.

\begin{table*}[htbp]
\caption{Number of knowledge triples in generated DKGs.}
  \label{tab:KG triple count}
  \Description{description}
\begin{tabular}{ccccc}
\hline
Dataset        & \begin{tabular}[c]{@{}c@{}}WTS\\ (GPT-3.5 w/CM)\end{tabular} & \begin{tabular}[c]{@{}c@{}}WTS\\ (GPT-3.5 w/EM-ESR)\end{tabular} & \begin{tabular}[c]{@{}c@{}}WTS\\ (GPT-3.5 w/EM-QSR)\end{tabular} & \begin{tabular}[c]{@{}c@{}}WYS\\ (GPT-4o w/EM-QSR)\end{tabular} \\ \hline
ChatDoctor5k    & 3,712                                                        & 3,554                                                            & 3,178                                                            & 3,136                                                           \\
PubMedQA        & 1,227                                                        & 1,635                                                            & 1,480                                                            & 1,178                                                           \\
MedMCQA(Single) & 1,024                                                        & 1,033                                                            & 1,124                                                            & 533                                                             \\
MedMCQA(Multi)  & 807                                                          & 1,361                                                            & 1,204                                                            & 655                                                             \\
SimpleQA        & 1,095                                                        & 1,134                                                            & 1,097                                                            & 828                                                             \\
sciq            & 181                                                          & 184                                                              & 191                                                              & 52                                                              \\
ScienceQA-NAT   & 156                                                          & 171                                                              & 146                                                              & 21                                                              \\
ScienceQA-SOC   & 43                                                           & 29                                                               & 25                                                               & 1                                                               \\
ScienceQA-LAN   & 385                                                          & 329                                                              & 317                                                              & 159                                                             \\ \hline
\end{tabular}
\end{table*}

\section{Prompts of Methods}

Taking MedMCQA, a medical QA dataset with choices as an example, we show the prompt used in the experiments of WTS, ChatGPT, CoT and ToG methods.

\subsection{WTS (Way-to-Specialist)}
\label{Prompts-WTS}

In WTS, there are four kinds of interaction with the LLMs. The following shows the prompts of WTS including Question Entity Extract Prompt, Triple Score and Prune Prompt, LLM Reason with Triples Prompt, and LLM Generate KG Triple Prompt.

As shown in Prompt \ref{WTS_Question_Entity_Extraction}, the "Question Entity Extraction" prompt of WTS  is designed to extract key entities from questions. The prompt instructs the LLM to function as a medical assistant, performing Named Entity Recognition to identify and output up to five meaningful entities. This prompt ensures that the most relevant information is captured concisely.

\begin{tcolorbox}[prompt={\begin{prompt} \label{WTS_Question_Entity_Extraction} Question Entity Extraction\end{prompt}}]

$\bullet$ system prompt: 

You are a [medical] assistant to carry out [Medical] Name Entity Recognition from a question and output JSON with only one key entity, like \{entities: [entity1, entity2,...]\}

 $\bullet$ user prompt: 
 
 Get at most 5 meaningful entities in the question: \{  Question \}.
\end{tcolorbox}

As shown in Prompt \ref{WTS_Triple_Score_and_Prune}, the "Score and Prune" prompt of WTS is designed to evaluate and rank the retrieved KG triples based on their relevance to a given question. The prompt instructs the LLM to assign a relevance score to each triple, and output a structured JSON object listing the most relevant triples in descending order of their scores.

\begin{tcolorbox}[prompt={\begin{prompt} \label{WTS_Triple_Score_and_Prune}Triple Score and Prune\end{prompt}}]

$\bullet$ system prompt: 

You are a [medical] expert to score the triples based on the relevance of the question and output JSON with only one key triple like \{ triples: [\{ triple: \{ head: xxx, relation: xxx, tail: xxx \}, score: xxx \},...] \}. The score is between 0 and 1, and the order is from high to low.

$\bullet$ user prompt: 

Based on the question: \{  Question \}, score the triples: \{ Triples \}.
\end{tcolorbox}

The "LLM Reason with Triples" prompt of WTS is designed to instruct LLM in generating answers to questions by reasoning with both its inherent knowledge and provided KG triples. The system prompt restricts the LLM to output a JSON object that includes the confidence level of LLM, the answer to the question, and the supporting information used to generate the answer, ensuring clarity in the response and precision in further evaluation.

\begin{tcolorbox}[prompt=
{\begin{prompt} \label{WTS_Reason_with_Triples}LLM Reason with Triples
\end{prompt}}]
$\bullet$ system prompt: 

You are a [medical] expert to answer patients' questions based on your own knowledge and the given information and give confidence [Yes/No]. The output should be a JSON with three keys confidence, answer, and support\_info like \{confidence: xxx, answer: xxx, support\_info: xxx\}

$\bullet$ user prompt: 

Based on your own knowledge and the <knowledge triple>, choose one of the <Option> to answer the question. knowledge triple: \{ Triples \},  Q: \{ Question \}, Option: \{ Option \}, A: ?
\end{tcolorbox}

The "Generate KG Triple" prompt in the WTS framework instructs the LLM to extract and construct KG triples from the provided question, answer, and entities. The prompt instructs the LLM to produce KG triples formatted with a head, relation, and tail, which creates a structured representation of knowledge.

\begin{tcolorbox}[prompt={\begin{prompt} \label{WTS_Generate_KG_Triple}LLM Generate KG Triple
\end{prompt}}]

$\bullet$ system prompt: 

You are a [medical] expert to extract general [medical] knowledge from question and answer to create [medical] knowledge graph triples. The output should be a JSON with only one key: triples like this: \{ triples: [<triple1>, <triple2>, ...] \}, and each triple should be like: \{ head: xxx, relation: xxx, tail: xxx \}.

$\bullet$ user prompt: 

Based on the question: \{  Question \},  answer: \{ Answer \} and entity: \{ Entity \}, extract and output triples.
\end{tcolorbox}

\subsection{ChatGPT}
\label{prompt-chatgpt}

For ChatGPT, we prompt it to directly generate the answer to the given question with one shot. The single example given is to assist its comprehension. To achieve accurate evaluations, we constrain the form of the output as JSON. The prompt is shown as Prompt \ref{ChatGPT_Directly_Generate_Answer}.

\begin{tcolorbox}[prompt={\begin{prompt} \label{ChatGPT_Directly_Generate_Answer}Directly Generate Answer
\end{prompt}}]
		
$\bullet$ system prompt: 

You are an assistant to answer questions to output JSON.
 \newline
 $\bullet$ user prompt: 
 
 Here is a question, choose one of the <Option> to answer the question. Your answer must be [0/1/2/3] which is the order of your choice and your output has only one key answer. 
 
 Q: \{ Question \}, Option:\{ Option \}, A: ?

Example:
        
Q: Which time zone is sesto ed uniti located in? 

Option: [Central European Time Zone, Greenwich Mean Time, Coordinated Universal Time, Central Standard Time] 

A: \{ answer: 0 \}
\end{tcolorbox}

\subsection{CoT (Chain-of-Thought)}
\label{prompt-CoT}
As shown in Prompt \ref{CoT_Directly_Generate_Answer}, with 5-shot prompting, the CoT prompt in our experiment aims to minimize modifications to the original text. To avoid any influence of the output format on the results, the output format is not constrained. 

\begin{tcolorbox}[prompt={\begin{prompt} \label{CoT_Directly_Generate_Answer}Directly Generate Answer
\end{prompt}}]
		
$\bullet$ system prompt:

You are an assistant to answer questions.

$\bullet$ user prompt: 

Select the [0/1/2/3] which is the order of your choice in Option as part of your answer. 

Example: 

Q: Sammy wanted to go to where the people were. Where might he go?  

Option: [race track, populated areas, desert, apartment] 
             
A: The answer must be a place with a lot of people. Race tracks, deserts, apartments, and roadblocks don't have a lot of people, but populated areas do. So the answer is (1).

Q: Google Maps and other highway and street GPS services have replaced what? 

Option: [united states, mexico, countryside, atlas] 

A: The answer must be something that is used to do what Google Maps and GPS services do, which is to give directions. Of the above choices, only atlases are used to give directions. So the answer is (3).
        
Q: Before getting a divorce, what did the wife feel who was doing all the work? 

Option: [harder, anguish, bitterness, tears]
        
A: The answer should be the feeling of someone getting divorced who was doing all the work. Of the above choices, the closest feeling is bitterness. So the answer is (2).
        
Q: What home entertainment equipment requires cable? 

Option: [television, radio shack, substation,  cabinet]
        
A: The answer must require cable. Of the above choices, only television requires cable. So the answer is (0).

Q: What kitchen appliance uses water to function? 

Option: [microwave, toaster, dishwasher, blender ]

A: The answer must use water to function. Of the above choices, only the dishwasher uses water to function. So the answer is (2).

Q: \{ Question \}, Option: \{ Option \}, A: ?
\end{tcolorbox}

\subsection{ToG (Think-on-Graph)}
\label{prompt-ToG}

The prompts for ToG are adapted from \cite{sun2023think} where ToG is proposed and tailored to the dataset used in this paper. In \cite{sun2023think}, the entities in the questions are provided in their datasets. Therefore, we incorporated a Question Entity Extraction module similar to that in WTS.

\begin{tcolorbox}[prompt={\begin{prompt} \label{ToG_Directly_Generate_Answer}Directly Generate Answer
\end{prompt}}]
		
$\bullet$ system prompt: 

You are an AI assistant that helps people find information.

$\bullet$ user prompt: 

Select the [0/1/2/3] which is the order of your choice in Option as part of your answer. 

Example: 

Q: Sammy wanted to go to where the people were. Where might he go?  

Option: [race track, populated areas, desert, apartment] 

A: The answer must be a place with a lot of people. Race tracks, deserts, apartments, and roadblocks don't have a lot of people, but populated areas do. So the answer is (1).
        
Q: Google Maps and other highway and street GPS services have replaced what? 

Option: [united states, mexico, countryside, atlas] 

A: The answer must be something that is used to do what Google Maps and GPS services do, which is to give directions. Of the above choices, only atlases are used to give directions. So the answer is (3).
    
Q: Before getting a divorce, what did the wife feel who was doing all the work? 

Option: [harder, anguish, bitterness, tears]

A: The answer should be the feeling of someone getting divorced who was doing all the work. Of the above choices, the closest feeling is bitterness. So the answer is (2).
        
Q: What home entertainment equipment requires cable? 

Option: [television, radio shack, substation,  cabinet]

A: The answer must require cable. Of the above choices, only television requires cable. So the answer is (0).

Q: What kitchen appliance uses water to function? 

Option: [microwave, toaster, dishwasher, blender]

A: The answer must use water to function. Of the above choices, only the dishwasher uses water to function. So the answer is (2).

Q:\{ Question \}, Option: \{ Option \}, A: ?
\end{tcolorbox}

\begin{tcolorbox}[prompt={\begin{prompt} \label{ToG_Question_Entity_Extraction}Question Entity Extraction
\end{prompt}}]

$\bullet$ system prompt: 

You are an assistant to carry out Name Entity Recognition from a question and output JSON with only one key entity, like \{entities: [entity1, entity2,...]\}
 \newline
 $\bullet$ user prompt: 
 
 Get at most 5 meaningful entities in the question: \{  Question \}.
\end{tcolorbox}

\begin{tcolorbox}[prompt={\begin{prompt} \label{ToG_Extract_Relation}Extract Relation
\end{prompt}}]

$\bullet$ system prompt: 

You are an AI assistant that helps people find information.

$\bullet$ user prompt: 
 
Please retrieve \%s relations (separated by semicolon) that contribute to the question and rate their contribution on a scale from 0 to 1 (the sum of the scores of \%s relations is 1).
\end{tcolorbox}

\begin{tcolorbox}[prompt={\begin{prompt} \label{ToG_Extract_Entity}Extract Entity
\end{prompt}}]

$\bullet$ system prompt: 

You are an AI assistant that helps people find information.

 $\bullet$ user prompt: 
 
Get the meaningful entities in the question. The answer should be like this:

Q: Doctor, I've been experiencing a condition called cryptorchidism. My testicles have not descended properly into the scrotum. What medical tests do I need to take?
                  
A:{{entities: [cryptorchidism, testicles]}} in the form of a list. 

Q: \{ Question \}
\end{tcolorbox}

\begin{tcolorbox}[prompt={\begin{prompt} \label{ToG_Score_Entity_Candidates}Score Entity Candidates
\end{prompt}}]
	
$\bullet$ system prompt: 

You are an AI assistant that helps people find information.

$\bullet$ user prompt: 

Please score the entities' contribution to the question on a scale from 0 to 1 (the sum of the scores of all entities is 1).

Example:

Q: The movie featured Miley Cyrus and was produced by Tobin Armbrust?

Relation: film.producer.film

Entities: The Resident; So Undercover; Let Me In; Begin Again; The Quiet Ones; A Walk Among the Tombstones

Score: 0.0, 1.0, 0.0, 0.0, 0.0, 0.0
The movie that matches the given criteria is "So Undercover" with Miley Cyrus and produced by Tobin Armbrust. Therefore, the score for "So Undercover" would be 1, and the scores for all other entities would be 0.

Q: \{ Question \}
Relation: \{ Relation \}
Entities: \{ Entities \}
\end{tcolorbox}

\begin{tcolorbox}[prompt={\begin{prompt} \label{ToG_Reason_Answer}Reason Answer
\end{prompt}}]

$\bullet$ system prompt: 

You are an AI assistant that helps people find information.

$\bullet$ user prompt: 

Given a question and the associated retrieved knowledge graph triplets (entity, relation, entity), you are asked to answer the question with these triplets and your knowledge. You must select the [0/1/2/3] which is the order of your choice in Option as part of your answer.

Q: Find the person who said \"Taste cannot be controlled by law\", what did this person die from?

Option: [illness, earthquake, murder, accident]

Knowledge Triplets: Taste cannot be controlled by law., media\_common.quotation.author, Thomas Jefferson

A: Based on the given knowledge triplets, it's not sufficient to answer the entire question. The triplets only provide information about the person who said "Taste cannot be controlled by law," which is Thomas Jefferson. To answer the second part of the question, it's necessary to have additional knowledge about where Thomas Jefferson's dead. In my opinion, the answer to the question is (0).

Q: The artist nominated for The Long Winter lived where? 

Option: [De Smet, Denmark, Italy, Tokyo]

Knowledge Triplets: The Long Winter, book.written\_work.author, Laura Ingalls Wilder
Laura Ingalls Wilder, people.person.places\_lived, Unknown-Entity Unknown-Entity, people.place\_lived.location, De Smet

A: Based on the given knowledge triplets, the author of The Long Winter, Laura Ingalls Wilder, lived in De Smet. Therefore, the answer to the question is (0).

Q: Who is the coach of the team owned by Steve Bisciotti? Option: [Donald J. Trump, Vincent Thomas Lombardi, Joe Paterno, John Harbaugh]

Knowledge Triplets: Steve Bisciotti, sports.professional\_sports \_team.owner\_s, Baltimore Ravens
Steve Bisciotti, sports.sports\_team\_owner.teams\_owned, Baltimore Ravens
Steve Bisciotti, organization.organization\_founder.organizations\_founded, Allegis Group

A: Based on the given knowledge triplets, the coach of the team owned by Steve Bisciotti is not explicitly mentioned. However, it can be inferred that the team owned by Steve Bisciotti is the Baltimore Ravens, a professional sports team. Therefore, additional knowledge about the current coach of the Baltimore Ravens can be used to answer the question. In my opinion, the answer to the question is (3).

Q: Rift Valley Province is located in a nation that uses which form of currency? 

Option: [Afghan afghani, Japanese yen, Kenyan shilling, Euro]

Knowledge Triplets: Rift Valley Province, location. administrative\_division.country, Kenya
Rift Valley Province, location.location.geolocation, UnName\_Entity
Rift Valley Province, location.mailing\_address.state\_province\_region, UnName\_Entity
Kenya, location.country.currency\_used, Kenyan shilling

A: Based on the given knowledge triplets, Rift Valley Province is located in Kenya, which uses the Kenyan shilling as its currency. Therefore, the answer to the question is (2).

Q: The country with the National Anthem of Bolivia borders which nations? 

Option: [Uruguay, Brazil, Venezuela, Colombia]

Knowledge Triplets: National Anthem of Bolivia, government. national\_anthem\_of\_a\_country.anthem, UnName\_Entity
National Anthem of Bolivia, music.composition.composer, Leopoldo Benedetto Vincenti
National Anthem of Bolivia, music.composition.lyricist, José Ignacio de Sanjinés
UnName\_Entity, government.national\_anthem\_of\_a\_country.country, Bolivia
Bolivia, location.country.national\_anthem, UnName\_Entity

A: Based on the given knowledge triplets, we can infer that the National Anthem of Bolivia is the anthem of Bolivia. Therefore, the country with the National Anthem of Bolivia is Bolivia itself. However, the given knowledge triplets do not provide information about which nations border Bolivia. To answer this question, we need additional knowledge about the geography of Bolivia and its neighboring countries. In my opinion, the answer to the question is (1).

Q: \{ Question \}
\end{tcolorbox}

\begin{tcolorbox}[prompt={\begin{prompt} \label{ToG_Information_Evaluate}Information Evaluate
\end{prompt}}]
		
$\bullet$ system prompt: 

You are an AI assistant that helps people find information.

$\bullet$ user prompt: 

Given a question and the associated retrieved knowledge graph triplets (entity, relation, entity), you are asked to answer whether it's sufficient for you to answer the question with these triplets and your knowledge (Yes or No). Besides, you must select the [0/1/2/3] which is the order of your choice in Option as part of your answer.

Q: Find the person who said \"Taste cannot be controlled by law\", what did this person die from? 

Option: [illness, earthquake, murder, accident]

Knowledge Triplets: Taste cannot be controlled by law., media\_common.quotation.author, Thomas Jefferson

A: {No}. Based on the given knowledge triplets, it's not sufficient to answer the entire question. The triplets only provide information about the person who said "Taste cannot be controlled by law," which is Thomas Jefferson. To answer the second part of the question, it's necessary to have additional knowledge about where Thomas Jefferson's dead. In my opinion, the answer to the question is (0).

Q: The artist nominated for The Long Winter lived where? 

Option: [De Smet, Denmark, Italy, Tokyo]

Knowledge Triplets: The Long Winter, book.written\_work.author, Laura Ingalls Wilder
Laura Ingalls Wilder, people.person.places\_lived, Unknown-Entity
Unknown-Entity, people.place\_lived.location, De Smet

A: {Yes}. Based on the given knowledge triplets, the author of The Long Winter, Laura Ingalls Wilder, lived in De Smet. Therefore, the answer to the question is (0).

Q: Who is the coach of the team owned by Steve Bisciotti? 

Option: [Donald J. Trump, Vincent Thomas Lombardi, Joe Paterno, John Harbaugh]

Knowledge Triplets: Steve Bisciotti, sports.professional\_sports\_team.owner\_s, Baltimore Ravens
Steve Bisciotti, sports.sports\_team\_owner.teams\_owned, Baltimore Ravens
Steve Bisciotti, organization.organization\_founder.organizations\_founded, Allegis Group

A: {No}. Based on the given knowledge triplets, the coach of the team owned by Steve Bisciotti is not explicitly mentioned. However, it can be inferred that the team owned by Steve Bisciotti is the Baltimore Ravens, a professional sports team. Therefore, additional knowledge about the current coach of the Baltimore Ravens can be used to answer the question. In my opinion, the answer to the question is (3).

Q: Rift Valley Province is located in a nation that uses which form of currency? 

Option: [Afghan afghani, Japanese yen, Kenyan shilling, Euro]

Knowledge Triplets: Rift Valley Province, location.administrative\_division.country, Kenya
Rift Valley Province, location.location.geolocation, UnName\_Entity
Rift Valley Province, location.mailing\_address.state\_province\_region, UnName\_Entity
Kenya, location. country.currency\_used, Kenyan shilling

A: {Yes}. Based on the given knowledge triplets, Rift Valley Province is located in Kenya, which uses the Kenyan shilling as its currency. Therefore, the answer to the question is (2).

Q: The country with the National Anthem of Bolivia borders which nations? 

Option: [Uruguay, Brazil, Venezuela, Colombia]

Knowledge Triplets: National Anthem of Bolivia, government. national\_anthem\_of\_a\_country.anthem, UnName\_Entity
National Anthem of Bolivia, music.composition.composer, Leopoldo Benedetto Vincenti
National Anthem of Bolivia, music.composition.lyricist, José Ignacio de Sanjinés
UnName\_Entity, government.national\_anthem\_of\_a\_country.country, Bolivia
Bolivia, location.country.national\_anthem, UnName\_Entity

A: {No}. Based on the given knowledge triplets, we can infer that the National Anthem of Bolivia is the anthem of Bolivia. Therefore, the country with the National Anthem of Bolivia is Bolivia itself. However, the given knowledge triplets do not provide information about which nations border Bolivia. To answer this question, we need additional knowledge about the geography of Bolivia and its neighboring countries. In my opinion, the answer to the question is (1).

Q:\{ Question \}, Option: \{ Option \}, Knowledge Triplets: \{ Knowledge Triplets \}, A: ?
\end{tcolorbox}

\end{document}